%% file: main.tex
\documentclass[preprint]{elsarticle}
\usepackage[ruled]{algorithm2e}
\usepackage{endnotes}
\usepackage{pgfplots}
\pgfplotsset{compat=newest}

\usepackage{verbatim}

\usepackage{amsmath,amsfonts,amsthm,amssymb}
\def\setR{\mathbb{R}}

\def\td{\texttt{TD} }
\def\RL{\texttt{RL} }
\def\rl{\texttt{DeepRL} }
\def\S{\mathcal{S}_t}
\def\A{\mathcal{A}_t}
\def\R{\mathcal{R}}
\usepackage{footnote}

\usepackage{graphics}
\usepackage{graphicx}
\usepackage{adjustbox}
\newcommand{\red}[1]{\textcolor{red}{#1}}

\usepackage{comment}
\usetikzlibrary{er,positioning, calc, shapes, arrows, chains, fit, shapes.geometric, shapes.symbols}
\tikzstyle{decision} = [diamond, draw, fill=white!20, 
    text width=4.5em, text badly centered, node distance=3cm, inner sep=0pt]
\tikzstyle{block} = [rectangle, draw, fill=white!20, 
    text width=6em, text centered, rounded corners, minimum height=3.5em]
\tikzstyle{line} = [draw, -latex']
\tikzstyle{cloud} = [draw, ellipse,fill=white!20, node distance=3cm,
    minimum height=2em]
\tikzstyle{rail}=[very thick]
\begin{document}
\title{ Solving the Train Dispatching problem via \\  Deep Reinforcement Learning}


\newcommand{\VALE}[1]{\todo[inline, color=blue!40]{VALE: #1}}

\newcommand{\LEO}[1]{\todo[inline, color=red!40]{LEO: #1}}


\author[1,2]{Valerio Agasucci}
\ead{agasucci@diag.uniroma1.it}

\author[3]{Giorgio Grani\corref{cor1}
}
\ead{g.grani@uniroma1.it}

\author[2]{Leonardo Lamorgese}
\ead{leonardo.lamorgese@optrail.com}

\address[1]{DIAG, Sapienza University of Rome, Rome, Italy}
\address[2]{OPTRAIL, Rome, Italy}
\address[3]{Department of Statistical Sciences, Sapienza University of Rome, Rome, Italy}
\cortext[cor1]{Corresponding author}


\begin{abstract}
    Every day, railways experience disturbances and disruptions, both on the network and the fleet side, that affect the stability of rail traffic. Induced delays propagate through the network, which leads to a mismatch in demand and offer for goods and passengers, and, in turn, to a loss in service quality. In these cases, it is the duty of human traffic controllers, the so-called dispatchers, to do their best to minimize the impact on traffic. However, dispatchers inevitably have a limited depth of perception of the knock-on effect of their decisions, particularly how they affect areas of the network that are outside their direct control. In recent years, much work in Decision Science has been devoted to developing methods to solve the problem automatically and support the dispatchers in this challenging task. This paper investigates Machine Learning-based methods for tackling this problem, proposing two different Deep Q-Learning methods(Decentralized and Centralized). Numerical results show the superiority of these techniques respect to the classical linear Q-Learning based on matrices. Moreover the Centralized approach is compared with a MILP formulation showing interesting results. The experiments are inspired on data provided by a U.S. class 1 railroad. 
\end{abstract}


\maketitle





\section{Introduction}\label{sec:intro}
A railway system is a complex network of interconnected tracks, where train traffic is controlled via the signaling system by activating switches and signals. The management and operation of a railway system is the task of the Infrastructure Manager (IM).  A Train Operating Company (TOC) organizes its fleet to accommodate expected demands, maximizing revenue and coverage. In regions where for geographical and historical reasons the market is made up of predominantly freight-hauling companies, such as, e.g., North America and Australia, the IM and TOC are often the same entity. That is, rail operators are generally vertically integrated, owning and operating both the network and the fleet. In most countries however, the IM is a single public authority, which rents its network to TOCs to operate their train services. This is particularly common for systems that operate predominantly passenger traffic, like in Europe. Here, the IM interacts with the TOCs to establish a  plan for train traffic, the so-called {\em timetable}. This process, referred to as timetabling, takes place offline, generally every 3 to 12 months, and is a challenging and time-consuming task. Its main goal is to assign routes to trains and create a schedule that is conflict-free and, typically, presents some elements of periodicity. Despite a vast body of literature in this field of research, in practice timetables are to this day still in many cases hand-engineered by specialised personnel, basing their decisions largely on experience, regulation, safety measures, business rules to factor in the various requests expressed by the TOC (the latter more delicate and time-consuming in liberalised markets with multiple competing TOCs). 

In operating a railway system, the IM ideally attempts to adhere perfectly to the timetable. Unfortunately, as anyone who has ever taken a train will be familiar with, this seldom happens, as small disturbances, or in some cases serious disruptions, occur daily.
A train malfunction, switch or track failure, delays in the preparation of the train or in passengers embarking, and plenty of other issues may create train delays and ultimately affect the overall network, sometimes in unforeseen ways. In some cases small delays are recovered from simply driving trains faster, but, very often, online re-routing and re-scheduling decisions have to be taken to reduce delays and increase efficiency. 
In literature, this online decision making process is referred to as the Train Dispatching problem (\td), a real-time variant of the above mentioned Train Timetabling problem (known to be NP-hard \cite{caprara2002modeling}). The very little time admitted for computation (often only a few seconds) and the size and complexity of real-life instances makes this problem very challenging to solve also in practice. The real-time nature of the problem in particular largely limits the solution approaches that can be used effectively. Operationally, the task of dispatching trains is in the hands of the IM's traffic controllers, the {\em dispatchers}, each assigned to a specific portion of the network (a station, a junction, a part of a line, etc). To this day, dispatchers are generally provided with little or no decision support in this process, which makes it challenging for them to go beyond their local view of the network and take into account the knock-on effects of their decisions, especially in areas of the network that are outside their direct control. 

The Train Dispatching problem has sparked much research interest over the years, in particular in the Optimization community. 
Different models and solutions approaches have been proposed over the years: some exact methods (\cite{lamorgese2015exact}-\cite{narayanaswami2013modelling}), heuristic based on MILP formulation as \cite{yan2012mixed}-\cite{boccia2012solving}-\cite{boccia2013dispatching}-\cite{adenso1999line}, tabu search \cite{tornquist2005train}, genetic algorithms \cite{higgins1997heuristic},  classic greedy heuristic \cite{cai1998greedy}, and neighbourhood search (\cite{sama2017variable}).
Refer to \cite{dispatchingHandbookChapter} for a deeper insight on the Train Dispatching problem, and to the many surveys on the topic for an overview of these approaches (e.g. \cite{CaEtAl14}-\cite{CoMe15}-\cite{fang2015}- \cite{narayanaswami2011scheduling}-\cite{tornquist2006computer}). Recent developments in both Machine Learning and Optimization have led to the definition of new learning-based paradigms to solve hard problems. Our focus is on Reinforcement Learning, for which a lot of interesting results have been achieved so far. In particular, in the well known AlphaGo algorithm (see \cite{silver2017mastering}) the authors tackle the complex game of Go, developing an outstanding framework able to overcome the best human player. Several approaches have followed AlphaGo, like AlphaZero \cite{silver2017mastering2,silver2018general} and MuZero \cite{schrittwieser2019mastering}, increasing every time the degree of generalization possible. Recently in \cite{khalil2017learning}, this approach was extended to combinatorial problems with a more general range of possible applications. To achieve their results, the authors combined Deep Q-Learning with graph convolutional neural networks, which are a specialized class of models for graph-like structures. A very similar approach was followed in \cite{drori2020learning}.
\medskip

\paragraph{\bf{Our contribution}}
In this paper, \td is tackled by means of Deep Q-Learning on a single railway line. More specifically, two approaches are investigated: Decentralized and Centralized. In the former, each train can be seen as an independent agent with the ability to see only a part of the network, namely some tracks ahead/behind it and not beyond. This approach has the advantage that it can be easily generalized, but, on the other hand, may lack the depth of prediction useful to express network dynamics. The latter method takes as input the entire line and learns to deal with delay propagation. 
Moreover, we use a Graph Neural Network to estimate this delay reflecting the railway topology. Both methods generalize to different railway sizes.
From the standpoint of the application, these authors find the use of Deep Reinforcement Learning approaches (such as Deep Q-Learning) very promising, as they present the advantage of shifting the computational burden to the learning stage. While enumerative algorithms typical of Combinatorial Optimization and Constraint Programming have proven to be effective in several train dispatching contexts, scalability remains a daunting challenge. On the other hand, under the assumption that the model can be trained effectively, Deep RL could achieve a (quasi) real-time performance, unattainable with classical optimization approaches. Finally, we point out that a possible reason for the slow adoption of automatic dispatching software in the industry is the diversity of business rules and requirements for such software in different regions and markets. The step of adapting the software to a specific set of such rules and requirements could be accelerated (or indeed skipped) using a Deep RL-based approach, which builds its internal input-output representation based on provided data and requires virtually no knowledge of the rules themselves.

For all the above, we believe that it is worth pursuing the application of Deep RL techniques in the field of train dispatching. 
This article is not the first article proposing the use of RL for train dispatching, so we highlight the differences between these. Papers \cite{Indian}-\cite{Sloveni} use a linear Q-Learning approach to tackle the problem. In \cite{Inglesi}, the authors present an approach based on approximate dynamic programming. 
Recent papers \cite{ning2019deep},\cite{Giapponesi},\cite{wang2019timetable} introduce a Deep-network to predict the next action.  {In \cite{liao2021deep} Deep RL is used to solve the energy-aimed timetable rescheduling (find the optimal timetable minimizing the energy usage) in \cite{wang2021policy} an RL approach is exploited to reschedule the timetable of the high-speed railway line between Shangai and Beijing, while in \cite{wang2023cooperative} a multi-agent reinforcement learning is incorporated in the metro system. }
 Finally in \cite{ying2020actor}-\cite{ying2021adaptive} an Actor-Critic method is used to schedule the underground train service in London.
However, these approaches are quite limited in terms of the size of the instances that can be solved. Indeed, in \cite{ning2019deep},\cite{Giapponesi},\cite{wang2019timetable} the railway network considered in the experiments is limited to 7 or 8 stations and in \cite{ning2019deep},\cite{wang2019timetable} train traffic is considered only in one direction. In this paper we present an approach which is able to tackle larger instances (up to 29 stations in our experiments) and handle both traffic directions. Furthermore, we model other, important factors in the 
dispatching process, such as train length and other business rules described in section \ref{Train_dispatching_problem}, which instead are not all covered in the cited papers. { The train length has been tackled in other works among the ones cited before, but it appears to be a factor that increases significantly the computational complexity of the proposed procedures, where here is directly embedded.}  This results in a model that is more faithful to real-life requirements and an algorithm that can solve instances of some practical relevance (e.g. a regional line).
In addition, it may be worth noting that our experiments are carried out on data from the US railway network, which presents its own challenges respect to other regions in the world like the European, Chinese and Japanese ones. 
Finally, in the computational section we show how the Deep Q-Learning approaches here presented perform better than their linear counterpart, the matrix Q-learning approach proposed in \cite{Indian}.

\medskip

The paper is organized as follows: in Section \ref{sec:preliminaries}, basic concepts of Train Dispatching and Reinforcement Learning are introduced formally. In Section \ref{sec:GNN} Graph Neural Networks are introduced. In Section \ref{sec:algorithms}, the two algorithms are discussed. Specifically, in sub-section \ref{sec:decentralized}, the Decentralized approach is tackled, whereas in sub-section \ref{sec:centralized}, the Centralized approach is presented. In Section \ref{sec:numericalresults}, a numerical analysis is conducted to prove the effectiveness of Deep Reinforcement Learning approach for the train dispatching problem. Finally in Section \ref{sec:MIP} a comparison between our approach and a well-known Mixed-Integer Programming approach is provided.

\section{Preliminaries}
\label{sec:preliminaries}

Two basic ingredients characterize this paper: the Train Dispatching problem (\td) and (Deep) Reinforcement Learning (\rl). In the following, a short presentation of both, with the aim of being introductory and not comprehensive. 

\subsection{Train Dispatching problem}
\label{Train_dispatching_problem}

The fundamental elements of the Train Dispatching problem (\td) are trains, the railway network, and paths. Trains are of course the means of transportation in this system. In practice, different types of trains can operate at the same time on a network. Two of the most important attributes of a train for
dispatching decisions are its priority and its length. Priority represents the relative importance of a train respect to other trains, which generally reflects in the decisions taken by dispatchers to recover from delay. For example, cargo trains usually have a lower priority than passenger ones. Their length depends on various aspects, such as technology, market and nature of the service.

A railway network is composed of a set of tracks, switches and signals. A common way to represent this network is to see it as an alternating sequence of tracks and stations. A station is a logical entity in the network that comprises multiple tracks and switches where, typically, the majority of routing and scheduling actions take place. Tracks are physical connections between two stations. Trains traveling in opposite directions can never occupy a track simultaneously, while certain tracks allow this for trains travelling in the same direction. 

In terms of infrastructure, stations are effectively also a set of interconnected track segments. An interlocking route is a sequence of track segments that connects two signals. Different routes can be conflicting, that is, certain movements may not be allowed on such routes at the same time. The interlocking is precisely the signaling device that prevents these conflicts. Certain interlocking routes also include stopping points, which allow train activities. 
Some of these activities can be embarking passengers or loading goods. Generally each stopping point can be occupied by at most one train at the time.
A switch is a mechanical installation enabling trains to be guided from one track to the other. A switch can be occupied by one train at a time.  Finally, a path is the expected sequence of tracks and stations for a given train. The path is specified by a starting station, a set of boarding points, and the arrival station.

 The elements introduced above allow  to model real-life operations with a level of approximation that is sufficient for the purposes of this paper. 
 \medskip
 
 In a few words, \td is the problem of managing train traffic in real-time by taking scheduling and routing decisions in order to maximize system efficiency. Such dispatching decisions are subject to a number of constraints related to the signaling system, infrastructure capacity, business rules, train physics etc. In railway systems operated following an official timetable, the goal is typically to adhere to it as much as possible. When deviations from this plan occur, the dispatcher objective is to recover from these deviations, minimizing train delay.

A particularly pernicious situation is the occurrence of a {\em deadlock}. A group of trains are said to be in deadlock if none of them can move because of another train that is blocking the next track of its path. A deadlock is the result of lack of information (e.g. wrong train length assumptions) or, more often, induced by human error (erroneous dispatching decisions). A critical aspect of any automatic dispatching systems is that it avoid creating deadlocks at all costs.

\subsection{Reinforcement Learning}
 Reinforcement Learning (\RL) is an approximated version of Dynamic Programming, as stated in \cite{bertsekas2019reinforcement}. This paradigm learns, in the sense that the approximated function built takes into account statistical information obtainable from previous iterations of the same algorithm or external data. The term Deep Reinforcement Learning (\rl) usually refers to \RL where a deep neural network is used to build the approximation. We refer to \cite{Libro} and \cite{bertsekas2019reinforcement} for a comprehensive discussion on \RL. 

\RL is usually explained through agent and environment interaction. The agent is the part of the algorithm that learns and takes decisions. To do so, it analyzes the surrounding environment. Once it takes a decision, the environment reacts to this decision and the agent perceives the effect its action has produced. To understand if the action taken has been successful or not, the agent may receive a reward associated with the action and the new environment observed.

In the Decentralized approach in Section \ref{sec:decentralized}, the agent is the train and the 
environment the observable line {, i.e. the portion of line and trains reachable by the agent, more details in Section \ref{sec:decentralized}}, as opposed to the centralized approach, the agent act as a line coordinator, deciding for each train, and the environment is the entire line.

For the purposes of this paper, the reinforcement procedure will move forward following the rolling horizon of events, indexed by $t$. 
An event takes place every time there is a decision to make that may affect the global objective.
In the \RL vocabulary, the state represents the formal representation of the environment at a time step $t$, and it is formalized by the vector $s_t \in \S$, where $\S \subseteq \setR^m$ is the set of all possible states at time $t$. The action taken by the agent is $a_t \in \A$, where $\A\subseteq\setR^n$ is the set of all possible actions at time $t$. Once the agent observes $s_t$ and carries out $a_t$, the environment reacts by producing the new state $s_{t+1}\in \S$ and a reward $r_{t+1}\in\R$, where $\R \subseteq \setR$ is the set of all possible rewards. In this case, $\R$ is mono-dimensional, but in general, it could take the form of a vector, depending on the problem examined.
Figure \ref{fig:RL_scheme} shows a simplified flowchart of \RL.

\begin{figure}[!t]
    \centering
    \input{RL_scheme}
    \caption{The reinforcement learning framework}
    \label{fig:RL_scheme}
\end{figure}

Q-learning is a branch of \RL  that uses an action-value function (usually referred to as q-function) to identify the action to take. More formally, the q-function $Q(s_t, a_t)$ represents the reward that the system is expected to achieve by taking said action given the state. At each step $t$, the action that brings the highest reward is chosen, namely:

$$a_t = \arg\max_{a \in \A} Q(s_t, a)$$

Pseudo-code for a Q-learning algorithm is presented in \ref{alg:RL}.

\begin{algorithm}\label{alg:RL}\caption{Q-learning algorithm}
\footnotesize
	\DontPrintSemicolon
	{\bf Input:}   $\cal P$ a set of instances, ${\cal D} = \emptyset$ the memory, a loss function ${\cal L}(\hat{y},y)$, $\epsilon \in (0,1)$, $\gamma \in (0,1]$, \texttt{\#episodes}, \texttt{\#moves}\;
	{\bf Output:} the trained predictor $Q(\cdot,\cdot)$\;
	
	\For{ $k = 1, \dots,$\texttt{\#episodes}}{
	{\bf Sample} $P \in \cal P$\;
	{\bf Initialize episode} $t=0$, $s_0 = 0$\;
	   \For{$t= 0, \dots,$\texttt{\#moves}}{
		{\bf Select} an action
		 $a_t =\left\{\begin{array}{l}
		    \textit{Unif}\left\{\A \right\},  \text{ with probability } \epsilon \\
		    \arg\max_{a \in \A} Q(s_t, a),   \text{ with probability } 1-\epsilon
		\end{array}  \right.$ \;
		{\bf Observe } $s_{t+1}, r_{t+1} = \texttt{ENVIRONMENT}(s_t, a_t)$\;
		{\bf Store } $(s_t, a_t, y_{t+1})$ in the memory $\cal D$, where $y_{t+1} = \left\{ \begin{array}{l}
		    r_{t+1}, \text{ if } t+1=T \\
		    r_{t+1} + \gamma \arg\max_{a \in \A} Q(s_{t+1}, a), \text{ oth.} 
		\end{array}\right.$\;
		{\bf Sample} batch $(x, y) \subseteq \cal D$\;
		{\bf Learn } by making one step of stochastic gradient descent w.r.t. the loss ${\cal L}(Q(x), y)$\;
		\If{$t+1 = T$}{
		{\bf Break}\;}
		
	}}
\end{algorithm}

${\cal L}(\hat{y},y)$ is the loss function used to perform the training,
i.e. a measure of the error committed by the model that one wants to minimize.
Examples of loss functions are Mean Squared error, Cross-Entropy, Mean Absolute Error and so on, see \cite{wang2022comprehensive} for an overview on the topic.
$\epsilon \in (0,1)$ is the probability of choosing a random action, $T$ is the final state of the process (i.e. when it is not possible to move anymore), $\gamma \in (0,1]$ is the future discount, a hyper-parameter reflecting the fact that future rewards may be less important.
To make the estimation more consistent, usually, a replay mechanism is used, so that the agent interacts with the system for a few virtual steps before learning. This is suitable in situations where the environment can be efficiently manipulated or simulated in a way that the computational cost is only marginally affected. 

\section{Graph neural networks}
\label{sec:GNN}
Graph convolutional neural networks, introduced by \cite{scarselli2008graph}, are used to approximate the Q-value function. Differently from Deep-neural network, GNN exploits the graph input structure of the railway to learn the Q-value function. The name \textit{convolutional} derives from the fact that the information is aggregated, shared and propagated among nodes according to how they are connected. This process is known as message passing, since the information of each node is propagated to the others for a given number of steps.  {The outputs of message passing are the so-called embedding vectors (one embedding for each node of the graph). The embedding vector collects information not only provided by the node to which it refers, but it brings information by other nodes of the graph.} For richer overview of GNN, please refer to \cite{zhou2018graph}.

The input of the network is a graph $G(N,A)$ where $N$ is the set of vertices and $A$ the set of arcs. At each node $j \in N$ is associated a vector of features $x_j \in \mathbb{R}^n$ where $n$ is the number of features. 
In our message passing architecture,  we use the scheme adopted by \cite{almasan2019deep}. Given an input graph, for each node $i$ we define the set of neighbours $\delta(i)^- = \left\{j \in V : (j,i) \in E\right\}$.

The first step of the GNN is to apply a mean function to all the neighbour features of each node $i$:
\begin{equation}
\label{eq:mean}
    m_i = \frac{\sum_{j \in ne_i} x_j}{|ne_i|},\ \ i \in V
\end{equation}
 {Where $m_i$ is the mean of the features of the neighbours of node $i$. In Figure \ref{fig:Graph} we show an example of a graph, the features and the mean (computed as reported above) associated to each node.} Then a concatenation is performed between $m_i$ and $x_i$, and put as input  of a feed-forward neural network $ReLU1)$ with $ReLU$ (see \cite{sharma2017activation}) as an activation function:
$$ l_i = \textit{ReLU} \left( W_1^\intercal \textit{concat}\left( x_i, m_i\right) + b_1 \right),\ \ i \in V$$
 {The term $l_i$ is the output, $W_1$ the weights and $b_1$ the biases of the  neural network $ReLU1$.}

Depending on the node connections the outputs are passed to a second neural network $ReLU2$ with $ReLU$ as an activation function.
$$ h_i = \textit{ReLU} \left( W_2^\intercal \left(l_i + \sum_{j \in ne_i} l_j \right) + b_2 \right),\ \ i \in V$$
Where $h_i$ is the embedding  {vector generated as the output of the neural network $ReLU2$, having $W_2$ and $b_2$ as weights and biases, respectively. The message passing returns an $h_i$ for each $i\in N$}. 
All the steps in the message passing can be performed several times, however for most applications 2 or 3 times are sufficient. In our experiments, we do so twice. The message passing ensures that the information of each node is propagated not only to its neighbours, but also to the furthest nodes. $ReLU1$ and $ReLU2$ have the same category of parameters: $W_1$ and $b_1$ for $ReLU1$; $W_2$ and $b2$ for $ReLU2$. The embedding vector $h_i$ may have a different size than $x_i$. This is justified by the fact that $h_i$ does not only bring the information from $x_i$ but also from other nodes. To have another message passing step $h_i$ becomes the new input of (\ref{eq:mean}).
Finally to obtain the target (in this work the Q-values) a last network $f (\cdot)$ is applied to the sum of $h_i$. 
$$
    o = f\left(\left( h_i\right)_{i \in V} \right)
$$
 {In Figure \ref{fig:Message_Passing}, we show the network described above referred to graph in Figure \ref{fig:Graph}.}
\begin{figure}
    \centering
    \input{grafo}
    \caption{Graph and its features.}
    \label{fig:Graph}
\end{figure}

\begin{figure}
    \centering
    \input{network_def}
    \caption{Message passing on the graph in figure \ref{fig:Graph}.}
    \label{fig:Message_Passing}
\end{figure}

\section{Proposed methods}

\label{sec:algorithms}
In this section, basic ideas regarding states, actions, and rewards to solve \td are presented. Two approaches are proposed: Decentralized and Centralized. The difference between them lies mainly in the topology of the state  and in the reward mechanism, and therefore in the ability of the approximating q-function to capture the right policy.
When describing the state, both of the approaches use the concept of resource. A resource is a track or a stopping point that can be occupied by one or more trains. 
\medskip

\medskip

The majority of models for \td rely on several approximations to make the problem more readable and the mathematics easier to handle. Using \RL allows  to integrate many of these hidden aspects in the environment, introducing a new level of complexity with a relatively small effort.
One of the most important is train length,  {which translate into computational burden for traditional optimization algorithms, where here is a feature directly embedded in the system. To avoid this,} some models in the literature tend to approximate a train with a point, but this is not the case in real life where a train may occupy more than one resource at the same time. This is especially critical in railway systems that operate predominantly freight services (such as in the North American market), where trains are often longer than the available infrastructure to accommodate them, and the risk of dispatcher-induced deadlocks is very real. 

Another important step is to model safety rules like the safety distance inside the same track, or the role of switches, 
For each track, we check the safety distance 
 between two consecutive trains. Additionally, we model the rule that if a switch is occupied by a train, then no other train can use the same switch. Therefore,  the  train that has to cross  will be held until the switch is  freed.
Finally,  we introduce the minimum headway time between the occupation of a track by consecutive trains. Given a couple of follower trains, we compute   the elapsed time since the first train has entered. If this quantity  is smaller than a certain threshold, the headway time, the train has to wait. In our experiments, the headway time, the safety distance and all the other railway parameters are exogenous value inspired by the US class 1 Railroad we are working with.
A simple rule has been implemented to avoid deadlocks between two crossing trains in specific circumstances.
In short, given a generic train $T_0$ positioned in $R_0$ that wishes to occupy the resource $R_1$, then the method evaluates the position of all visible trains $T_i$ and the single-track resources\footnote{A single-track resource is a resource that does not have a parallel resource} $R_i$, for $i=1,\dots,K$\footnote{$R_i$ is directly connected with $R_{i+1}$ $\forall i = 1, \dots K$}, converging to $R_1$. For each train $T_i$, the flow of resources ${\cal F}_i$ from $R_i$ to $R_1$ is then computed. Finally, if there exists at least one $i\in \{1,\dots,K \}$ such that a train $T_i$ occupy a resource $R_i$ in ${\cal F}_i$, then $T_0$ must wait to avoid deadlock.

 The inclusion of  all these aspects allows a greater adherence to the real-world problem than many optimization approaches attain, and
the relative ease in doing so is, in our opinion, one of the advantages of a deep \RL approach.


\subsection{Decentralized approach}
\label{sec:decentralized}

In this subsection,  we will present the Decentralized approach, introducing the structure of   the state  when full observability of the rail network is guaranteed. 

As studied in \cite{chen2018scalable,yang2019sample}, enriching the information available to the agent (so the state) affects the space of policies to be learned. For this reason, the following six features are associated to each resource:
\begin{enumerate}
     \item {\bf status}, which is a discrete value chosen among stopping point, track, blocked\footnote{Temporarily reserved by another train.}, and failure\footnote{The resource is out of service.}
    \item {\bf number of trains} 
    \item {\bf train priority} (if there is more than one train, the train with the highest priority is reported)
    \item {\bf direction}, which is a discrete value chosen among: follower, crossing, or empty\footnote{A train T1 is a follower  for a train T2 if they have the same direction, otherwise T1 is a crossing train for T2.}
    \item {\bf length check}, a Boolean identifying if the train length is less than or equal to the resource size
    \item {\bf number of parallel resources}\footnote{Given a resource $R_1$ that connect two resources $R_A$ and $R_B$, a resource $R_2$ is a parallel resource of $R_1$ if  $R_2$ connects $R_A$ and $R_B$. } w.r.t. to the current one
\end{enumerate}
\medskip

For that which concerns the learning process the model that approximates the q-function is represented by a feed-forward deep neural network (\texttt{FNN}) with two fully connected hidden layers of 60 neurons each, and a third layer mapping into the space of the actions. The output of the third layer is then combined with a mask, disabling infeasible moves. Figure \ref{fig:Network_model} shows the structure of the network.
Firstly, the input (the state) passes through a layer of neurons, where each neuron is associated with a ReLU (see \cite{sharma2017activation}) activation function. The output of the first layer   goes to a second one with the same kind of activation functions.  Then, everything is multiplied by the action mask, filtering allowed actions from prohibited ones. An action mask is a vector, whose components are equal to one if the correspondent action is allowed, zero otherwise.

\begin{figure}[!t]
    \centering
    \input{network_model}
    \caption{The deep network utilized}
    \label{fig:Network_model}
\end{figure}


The typical goal of a dispatcher is to take routing and scheduling decisions that minimize some measures of delay. One way to model these decisions is to establish whether a train can access a specific resource or whether it has to stop/hold before entering it. 
The algorithm takes this decision when a train reaches a control point. When possible, the train will take the best-programmed resource as default. This happens when, for instance, all the resources ahead of the train are free. The best resource is the one ensuring a minimum programmed running-time for the specific train.

In Figure \ref{fig:Action_description}
, for example,  we have four resources: 
1a, 1b. 1c and 1d with respectively 800, 1000, 1200 and 1300 unimpeded running time,  {which is is the minimum running time that a train needs to go from its current position to its destination (in this example 1a, 1b, 1c and 1d) if it were never held.}.
 The resource 1a represents the best choice, since it accumulates less unimpeded running time, while the alternatives  (1b, 1c, 1d) are all  higher.

On the other hand, it may happen that all the next reachable resources are not available, so the train is forced to stop. In all the other situations, the algorithm acts as if the train were able to take decisions by itself. The state of the system (from the point of view of the train) does not represent the entire network, but only a limited number of resources ahead and behind. For each visible resource, the six features discussed before are considered.
\begin{figure}
\centering
\begin{tikzpicture} 

\draw[-] (-5,9.5)--(-17,9.5);
\draw[-] (-10,10.25)--(-12,10.25);
\draw[-] (-12.5,9.5)--(-12,10.25);
\draw[-] (-9.5,9.5)--(-10,10.25);
\draw[-] (-12.5,11)--(-9.5,11);
\draw[-] (-9,9.5)--(-9.5,11);
\draw[-] (-9,9.5)--(-9.5,11);
\draw[-] (-13,9.5)--(-12.5,11);

\draw[-] (-13,11.75)--(-9,11.75);
\draw[-] (-8.5,9.5)--(-9,11.75);
\draw[-] (-13.5,9.5)--(-13,11.75);

\draw[-] (-16,10)--(-14,10);
\draw[-] (-14,10)--(-14.5,10.25);
\draw[-] (-14,10)--(-14.5,9.75);


\node[text width=0.5cm] at (-11,9.75) {1a};
\node[text width=0.5cm] at (-11,10.5) {1b};
\node[text width=0.5cm] at (-11,11.25) {1c};
\node[text width=0.5cm] at (-11,12) {1d};
\node[text width=0.5cm] at (-15.25,10.25) {Train};


\end{tikzpicture}
    
    \caption{Best and reachable resources}
    \label{fig:Action_description}
\end{figure}

Given the state, the action to be taken is one of the following:
\begin{itemize}
    \item halt
    \item go to the best resource
    \item go to a reachable resource 
\end{itemize}

The first action (stopping a train) can always be taken, while the other two possibilities depend on the state.
\medskip

The most delicate part in the Decentralized approach is the definition of the reward. While the global objective is to minimize the total weighted delay, the agent viewpoint does not allow to express the reward associated with a state-action pair in terms of the actual impact on the overall network objective. For this reason, the reward is assigned at the end of each episode, and then the data collected is stored as $(s_t,a_t,Q(s_t,a_t),r_t)$, where $Q(s_t,a_t)$ is the output of the neural network.  In particular, the reward for each generated state-action pair is given a large penalty value if the episode ends up in a deadlock, a minor penalty if the weighted delay is greater than 1.25 times the minimum weighted delay found so far, and a prize if the weighted delay is less than or equal to 1.25 times the minimum weighted delay found so far. This strategy is inspired by \cite{Indian}, where the authors adopt a classical matrix-based Q-learning approach on a single-track.
We define the weighted delay of a train as the difference between the actual running time and the unimpeded running time.  {The actual running time is the timing of the solution taken by our algorithm, which may be different from the planned one, whereas the unimpeded running time is the  the minimum running time that a train needs to go from its current position to its destination.} By  multiplying this value by a priority factor, leveraging  how disruptive a delay would be, obtaining the cumulative delay of the train considered.  {The choice of leveraging reflects the different priorities of trains, so that a delay has more serious consequences if it refers to passenger train rather than a freight one. This is, in our experience, a common choice for companies.} Summing up all the single delays we obtain the total weighted delay.

\medskip

The last aspect to be discussed is memory management. Since no prior knowledge is used, the composition of the sample is critical to drive a smooth learning process. Therefore, the memory was divided into three data-sets: {\em best}, \emph{normal} and \emph{deadlock}. 
The \emph{best} memory stores all the samples ending up in a weighted delay that is less than or equal to 1.25 times the best delay found so far, the {\em normal} memory stores the ones with weighted delay that is greater than 1.25 the best found, and the \emph{deadlock} memory stores all the unsuccessful instances.

Action-state-reward items are stored according to the level of reward if and only if the action mask in the \texttt{FNN} allows more than one action, so as to strengthen the learning process only on critical moves.
 Deadlock and normal memory data-sets are never deleted, while the best memory is updated every time a new best weighted delay is found. 

\subsection{Centralized approach}
\label{sec:centralized}
The idea behind the Centralized approach is that the \rl algorithm may take advantage of knowing the state of the overall network at each step, and attempt to learn the particular dynamics of the network thanks to the GNN. In this case the agent can be seen as a line coordinator, deciding critical issues at control points, predicting the expected effect on the network. This mimics to some extent the behaviour of human dispatchers, which have full control over a limited part of the network. Moreover, considering all the network as a state, the use of a GNN instead of a feed-forward neural network allows adopt the  same neural network for  different railway sizes. In feed-forward networks, the input is fixed in size, requiring a different model for every railway case, whereas using GNNs the same model operates to multiple railway networks.


In this approach the state is a graph, each node is a resource (track or stopping point) and there is an arc when two resources are linked. With each node is associated a vector of features that explains the characteristics of the train in that resource. This is a one hot-encoded vector with the following characteristics:
\begin{itemize}
    \item the first $n$ bits are reserved for the train's priority. The $i$-th bit is set to  $1$, if the train belongs to the $i$-th priority class, and $0$ otherwise.  
    \item one bit at position $n+1$, for the train's direction in that resource
    \item $m$ bits, from position $n+2$ to position $n+2+m$, are reserved for the class length of the train
    \item finally, one bit in position $n+2+m+1$ states if a decision has to be taken for the train in that resource.
\end{itemize}


The action space is the same as the Decentralized approach, however the reward mechanism, unlike the Decentralized approach, takes into account the actual delay in absolute value, rather than a measure of how good such delay is with respect to the best achievable.
More in detail we define the reward-to-go from a time step $t$ as:
$$R_t = -\sum_{k=t}^T r_k$$
Where $r_k$ is the difference of delay between step $k$ and step $k-1$. The reward-to-go is a negative quantity because we are minimizing  the delay.
The Q-function, instead, given a state $s_t$ and the action $a_t$ is given by:
$$Q(s,a) = \mathbb{E}[\sum_{k=t}^T r_k|s=s_t,a=a_t]$$

To estimate the reward-to-go and measure the goodness of an action, we have therefore to minimize the following loss:
$$\mathbb{L} = \sum_{t=1}^T(Q(s_t,a_t)-R_t)^2.$$


\section{Numerical results}
\label{sec:numericalresults}

The test bed for the experiments is inspired on data provided by a U.S. class 1 railroad. 
The two algorithms were compared to the linear Q-learning algorithm proposed in \cite{Indian}, since the primary aim of this paper is to show that \rl is superior to linear Q-learning. In particular, the quality of the solutions in terms of final delay is the main driver used to discriminate the quality of an algorithm. For what concerns time, all the procedures need less than one second to complete an episode, and therefore are well suited for this online/real-time application.

All tests were performed on an Intel i5 processor, with no use of GPUs. 
Our experiments take into account railway networks with a different number of resources, number of trains and train's priorities.

The experiments were conducted to investigate the following properties:
\begin{enumerate}
    \item the ability to solve unseen instances generated from the same distribution
    \item the ability to generalize when the number and the length of the trains increase
    \item the ability to generalize when the numbers of resources in the railway network changes.
\end{enumerate}

\subsection{First experiment}
In this experiment the considered railway network is mainly characterized by the alternation of one station and one track (single-track), but also include some parts of the network where two stations are connected by two single tracks or more (multiple-track).
In total, the network has $134$ resources (tracks and stopping points), $15$ tracks and $29$ stopping points have at least one parallel resources, while $33$ are single-track resources.  {The time window is two hours.}
In Figure 6, 
we report a small section of our railway to show what we consider as a stopping point and as a track. Resources 2a and 2b are stopping points where trains can meet or pass. Elements 1, 3, 4a, 4b, 5a and 5b are tracks, where the couples 4a, 4b and 5a, 5b are parallel, meaning they are connected to the same stations. We refer to both tracks and stopping points as resources. The track is effectively the portion of the network that connects two stations. Stations can have stopping points or they can be crossovers, like $C1$ or $C2$, where a crossover is composed by one or more switches connecting two parallel tracks, or a parallel track and a single one. We do not model the switches, considering them as a part of the next track. As explained in Section \ref{sec:algorithms}, if a train occupies a switch no other train can use it.

\begin{figure}
\centering
\begin{tikzpicture} 

\draw[-] (-8,9.5)--(-10,9.5);
\draw[-] (-8,8.5)--(-10,8.5);
\draw[-] (-10,9.5)--(-12,9.5);
\draw[-] (-10,8.5)--(-12,8.5);
\draw[-] (-12,8.5)--(-14,8.5);
\draw[-] (-12,9.5)--(-12.5,8.5);
\draw[-] (-14,8.5)--(-16,8.5);
\draw[-] (-14,8.5)--(-14.5,9.5);
\draw[-] (-14.5,9.5)--(-15.5,9.5);
\draw[-] (-15.5,9.5)--(-16,8.5);
\draw[-] (-16,8.5)--(-18,8.5);

\draw[-] (-9.5,9.5)--(-10.5,8.5);
\draw[-] (-10.5,9.5)--(-9.5,8.5);

\node[text width=0.5cm] at (-17,8.3) {1};
\node[text width=0.5cm] at (-15,8.3) {2a};
\node[text width=0.5cm] at (-15,9.8) {2b};
\node[text width=0.5cm] at (-13,8.3) {3};
\node[text width=0.5cm] at (-11.5,8.3) {4a};
\node[text width=0.5cm] at (-8.5,8.3) {5a};
\node[text width=0.5cm] at (-11.5,9.8) {4b};
\node[text width=0.5cm] at (-8.5,9.8) {5b};
\node[text width=0.5cm] at (-12.5,9.15) {C1};
\node[text width=0.5cm] at (-10,9.8) {C2};


\end{tikzpicture}
    \label{fig:Rail_example}
    \caption{Railway example}
\end{figure}

Traffic characteristics for each instance are described in terms of:
number of trains, 
position,
direction, 
priority and
length.

The range for each parameter is realistic, as again it is inspired on input provided by the U.S. class 1 railroad. More specifically:
\begin{itemize}
\item the number of trains $N$ is chosen randomly between 4 and 10 with the following probability distribution: $P(N=4)=0.1$, $P(N=5)=0.2$, $P(N=6)=0.2$, $P(N=7)=0.2$, $P(N=8)=0.15$, $P(N=9)=0.1$, $P(N=10)=0.05$
\item the  position is chosen using a  uniform distribution on the available resources
\item the  direction is chosen uniformly 
\item the  priority $A$ is chosen randomly between 1 and 5 with the following probability distribution
$P(A=1)=0.05$, $P(A=2)=0.15$, $P(A=3)=0.23$, $P(A=4)=0.27$, $P(A=5)=0.3$
\item the length is chosen uniformly in the set \{4000, 4500, 5000, 5500, 6000, 6500\} to be intended in feet
\end{itemize}

 {
Given the above mentioned distributions, to generate an instance we  repeat the following steps:
\begin{enumerate}
    \item select the number of trains $N$ 
    \item for each train in $N$:
    \begin{enumerate}
        \item select its initial position  
        \item select its direction (down-hill, up-hill) 
    \item select its priority 
    \item select its length.
    \end{enumerate}
\end{enumerate}}
Priority, direction and position are sampled according to the distribution probability described above even in the next experiments.
Additionally, the delay is multiplied by a penalty factor to express the relative importance of a train class and its priority. Given a priority $A \in \left\{1,2,3,4,5\right\}$, the coefficient $\omega_A$ is such that $\omega_1 = 20$,  $\omega_2 = 10$,  $\omega_3 = 5$,  $\omega_4 = 2$ and $\omega_5 = 1$. 
Delays are affected by a penalty factor higher than one, which is linked to the priority. Since we are reporting delays multiplied by this factor, their values may appear high.

\medskip

The models were trained on $100$ randomly generated instances, with the fixed rail network described above, and tested on $100$ unseen instances with the same specifications.
The training phase involved running $10\,000$ episodes for each algorithm. An episode is stopped either when all trains are at their last resource or after a 2-hour plan is produced. 
In this context, we define an episode as the production of a 2-hour plan.
 {In Table \ref{tab:Stat_100} we show some statistics, like minimum, average and maximum delay, its standard deviation and the number of deadlocks.In  Table \ref{tab:WDL_G}, we report the number of instances won, drawn, and lost by each approach (Centralized, Decentralized and Q-Learning). We have a win when the delay found is smaller, a draw when it is comparable, and a loss it is higher with respect to the values obtained by the other approach in comparison. By looking at the results, we see the Centralized approach reaches a delay that is less than Decentralized in 44 instances, being comparable in 47, and higher in only 9 of them. Comparing Centralized and Q-Learning, we have that in 93 instances the Centralized reaches a delay that is less than Q-Learning, and similarly can be said for the Decentralized approach. This highlights the Centralized has the best performances on this test set.}
Figure \ref{fig:performance_profile} reports the performance profiles for the three algorithms w.r.t. the value of the delay obtained on 100 instances of the same distribution of the training set. 
Performance profiles have been used as proposed in \cite{DM2002}.
Given a set of solvers $\mathcal{I}$ and a set of problems $\mathcal{P}$, the performance profile takes as input a ratio between
 the performance  (i.e. the value of the weighted delay) of  a solver $i \in \mathcal{I}$ on problem $p \in \mathcal{P}$
and the best performance obtained by any solver in $\mathcal{I}$
on the same problem. Consider the cumulative 
function $\rho_s(\tau) = |\{p\in \mathcal{P}:\; r_{p,i}\leq \tau \}| /|\mathcal{P}|$ where $t_{p,i}$ is the delay and
$
r_{p,i} = t_{p,i}/\min\{t_{p,i^\prime}: i^\prime \in\mathcal{I}\}$.
The performance profile is the plot of the functions $\rho_s(\tau)$ for $s \in S$. Informally, the higher the curve the better the component. Here, the components are algorithms and their performance is determined by comparing the delays obtained on each instance. The graph shows the supremacy of deep architectures with respect to the simple linear one, which is reasonable due to the known complexity of the problem. In general, the Centralized model seems to perform better than the Decentralized one. 

 \begin{table}[!t]
		\centering
		\caption{Basic Statistics on $100$ Test Instances From the Same Distribution of the Training.}\label{tab:res}
		{
			\begin{tabular}{cccc}
			\hline  Delay & Linear Q-learning & Decentralized & Centralized  \\ \hline
			minimum & 886&	0&	0\\
average & 42399.04 &	9914.864 &	9295.297\\
maximum &174262 &	30578 &	35441 \\
std dev &35422.18 &	7287.825 &	7398.979\\
\# deadlocks& 4 & 5 & 4\\
			\hline 
		\end{tabular}}
	 \label{tab:Stat_100}
			\end{table}
  
  \begin{table}
    \centering
    \caption{ {Wins, draws and defeats on 100 network test instances.}}
    \begin{tabular}{cccc}
        
    \hline
        \textbf{Win/Draw/Loss} &\textbf{Centralized}&\textbf{Decentralized}&\textbf{Q-Learning}\\\hline
        Centralized & -&44-47-9&93-0-7 \\
        Decentralized &9-47-44&-&96-0-4\\
        Q-Learning&7-0-93 &4-0-96  &-\\ \hline
    \end{tabular}

    \label{tab:WDL_G}
\end{table}

\begin{figure}[!t]
\centering
    \input{SheetV1}
    \caption{Performance profiles for the three algorithms compared together based on the value of the delay obtained on $100$ instances of the same distribution of the training set.}
   \label{fig:performance_profile}
\end{figure}

Table \ref{tab:res} summarizes some basic performance statistics. As one can see, the average delay of the \rl approaches is considerably smaller, so \rl captures inner non-linearities more efficiently and appears more effective than linear Q-learning. 

\subsection{Second experiment}
As a second experiment, we test the ability of the three algorithms to generalize to longer trains with different frequencies. Longer trains in a network increase the probability to end up in a deadlock. In fact, long trains can occupy at the same time more than one resource and one or more switch. Even very simple configurations (e.g. one station with two stopping points and two crossing trains occupying both stopping points and switches) can lead to deadlocks. In other words, taking into account the length of trains introduces a whole new level of complexity, which is especially for freight-based traffic, where trains often tend to be very long.  {The time window considered is two hours.}

In particular, for the test set the new parameters adopted are:
\begin{itemize}
    \item Number of trains $N$; chosen randomly between 4 and 12 with the following probability distribution: $P(N=4)=0.05$, $P(N=5)=0.15$, $P(N=6)=0.15$, $P(N=7)=0.15$, $P(N=8)=0.15$, $P(N=9)=0.1$, $P(N=10)=0.1$, $P(N=11)=0.1$,$P(N=12)=0.05$
    \item Length; chosen uniformly in the set
    \{4000, 4500, 5000, 5500, 6000, 6500, 7000, 7500, 8000\}, 
to be intended in feet

\end{itemize}

The focus  is posed on these two parameters, since it has been observed empirically that they seem to be major factors in influencing the complexity of a \td problem.

In this case, the \rl Decentralized model is trained for $20\,000$ episodes, whereas the linear Q-learning for $50\,000$; we use the same learning of the previous experiment for the Centralized model. 

The Q-learning and Decentralized \rl were trained on $200$ instances from the distribution described in the first experiment  (except for length and number of trains) and tested on $200$ instances with the specifics specified above. Table \ref{tab:res_200} summarizes some basic performance statistics for this experiment.
Both Decentralized approach and Centralized outperforms Q-Learning, moreover Centralized finds almost half deadlock than Decentralized and as in the previous experiment the average delay in Centralized is less than Decentralized.
Linear Q-Learning appears to find less deadlocks than Decentralized approach. However, this is arguably induced by the fact that, in many test instances, the Q-learning based algorithm halts trains before they can reach a potential deadlock. This leads to huge delays, but falls short of explicitly creating a deadlock. In other words these results are somewhat biased, since in real-life a plan where trains are halted continuously (like those produced here by the linear Q-Learning approach) would be deemed equally unacceptable. 
\newline
 {A last observation on  the number of deadlocks, is that all the algorithms compared in Table \ref{tab:res_200} act as powered greedy heuristics. For this reason, their greedy nature may lead to unwanted solutions, which are more evident when the model cannot see the entire network, like in the Decentralized approach. We must not forget that the nature of the test instances is meant to consider also overcrowded situations, where a feasible solution may be hard to reach for a heuristic method. To the best of our experience, intensifying the training for longer periods or increasing the number of weights in the models may alleviate this circumstance.
}

 {Regarding the comparison among the different approaches, also in this case we appreciate the fact that both Centralized and Decentralized approaches outperforms the Q-learning. In Table \ref{tab:WDL_G_200}, we can see that in 107 test instances on 200 the Centralized approach returns a delay that is less than the Decentralized one while in 182 test instances returns a delay that is less than the Q-Learning.}
Figure \ref{fig:performance_profile2} shows the performance profiles in this case.

\begin{table}[!t]
	\centering
	\caption{Basic Statistics on $200$ Test Instances from the Same Distribution of the Training.}
	\label{tab:res_200}
			 {
			\begin{tabular}{cccc}
			\hline  Delay & Linear Q-learning & Decentralized & Centralized  \\ \hline
			minimum & 0&	0&	0\\
average & 50573.49 & 13966.1 &	13012.775\\
maximum & 256883.1 &	86309.01&	54979 \\
std dev &52318.19 &	14092.37 &	12045.295\\
\# deadlocks& 18 & 26 & 14\\
			\hline 
		\end{tabular}}

\end{table}
  \begin{table}
    \centering
    \caption{ {Wins, draws and defeats on $200$ Test Instances from the Same Distribution of the Training.}}
    \begin{tabular}{cccc}
        
    \hline
        \textbf{Win/Draw/Loss} &\textbf{Centralized}&\textbf{Decentralized}&\textbf{Q-Learning}\\\hline
        Centralized & -&107-70-23&182-3-15 \\
        Decentralized &23-70-107&-&189-3-8\\
        Q-Learning&15-3-182 &8-3-189  &-\\ \hline
    \end{tabular}

    \label{tab:WDL_G_200}
\end{table}

\begin{figure}[!t]
\centering
    \input{SheetV2}
    \caption{Performance profiles for the three algorithms compared together based on the value of the delay obtained on $100$ with higher number of trains (10 to 15) than the training set.}
   \label{fig:performance_profile2}
\end{figure}

\subsection{Third experiment: different sizes of network}

In the last experiments, we test the generalization capability when increasing the number of resources in our network. 
We define three different railways: $R1$, $R2$, $R3$,  {with the same time window of two hours.} Some characteristics of the railways are shown in Table \ref{tab:res_summary}.
\begin{table}[!t]
	\centering
	\caption{Railway network characteristics}
	\label{tab:res_summary}
			 {
			\begin{tabular}{ccccc}
			\hline  Railway & Total resources & Stopping points & Single track resources&Double tracks  \\ \hline
			$R1$ & 155&	79& 52	&12\\
$R2$ & 180 & 96 &60	&12\\
$R3$ & 196 &	106&	60 & 15 \\
			\hline 
		\end{tabular}}
\end{table}
Railway $R1$ has the following parameters in terms of train traffic characteristics:
\begin{itemize}
    \item Number of trains $N$; chosen randomly between 4 and 12 with the following probability distribution: $P(N=6)=0.05$, $P(N=7)=0.1$, $P(N=8)=0.1$, $P(N=9)=0.15$, $P(N=10)=0.15$, $P(N=11)=0.1$, $P(N=12)=0.1$, $P(N=13)=0.1$
    \item Length; chosen uniformly in the set
    \{4000, 4500, 5000, 5500, 6000, 6500, 7000, 7500, 8000\}
\end{itemize}
Railways $R2$ and $R3$ have the following parameters in terms of train traffic characteristics:
\begin{itemize}
    \item Number of trains $N$; chosen randomly between 4 and 12 with the following probability distribution: $P(N=7)=0.05$, $P(N=8)=0.1$, $P(N=9)=0.15$, $P(N=10)=0.15$, $P(N=11)=0.15$, $P(N=12)=0.15$, $P(N=13)=0.1$,$P(N=14)=0.1$, $P(N=15)=0.05$
    \item Length; chosen uniformly in the set
    \{4000, 4500, 5000, 5500, 6000, 6500, 7000, 7500, 8000\}
\end{itemize}
We generated 50 tests for each railway configuration.
We did not train our networks on these new infrastructures, and we rather use the one adopted in the
first experiment. In tables \ref{tab:res_155}-\ref{tab:res_175}-\ref{tab:res_196}, some statistics  for the three approaches. As shown in the tables, the two Deep-Learning approaches outperform linear Q-Learning, while the Centralized one again obtains better results in terms of weighted delay with respect to the Decentralized algorithm.  {In Tables \ref{tab:WDL_155}-\ref{tab:WDL_180}-\ref{tab:WDL_196} we can see how Centralized approach in about one third of the instances reach a delay that is less than the Decentralized,  and it is comparable for one fourth of them.}

\begin{table}[!t]
	\centering
	\caption{Basic Statistics on a new railway with $155$ resources}
	\label{tab:res_155}
			 {
			\begin{tabular}{cccc}
			\hline  Delay & Linear Q-learning & Decentralized & Centralized  \\ \hline
			minimum & 1572&	0&	0\\
average & 54611.55 & 16597.912 &	14074.175\\
maximum & 153198.146 &	51952.006&	47270.46 \\
std dev &43666.76 &	14144.157 &	12861.334\\
\# deadlocks& 5 & 5 & 5\\
			\hline 
		\end{tabular}}
	
\end{table}

\begin{table}[!t]
	\centering
	\renewcommand{\arraystretch}{1.3}
	\caption{Basic Statistics on a new railway with $175$ resources}
	\label{tab:res_175}
			 {
			\begin{tabular}{cccc}
			\hline  Delay & Linear Q-learning & Decentralized & Centralized  \\ \hline
			minimum & 5412&	1074&	392\\
average & 68347.693 & 27082.873 &	17157.326\\
maximum & 246231.062 &	112981.007&	52735.47 \\
std dev &48582.189 &	24898.419 &	14324.176\\
\# deadlocks& 4 & 11 & 14\\
			\hline 
		\end{tabular}}
	
\end{table}

\begin{table}[!t]
	\centering
	\caption{Basic Statistics on a new railway with $196$ resources}
	\label{tab:res_196}
			 {
			\begin{tabular}{cccc}
			\hline  Delay & Linear Q-learning & Decentralized & Centralized  \\ \hline
			minimum & 4846&	0&	0\\
average & 61033.829 & 21250.22 &	16844.557\\
maximum & 159952.036 &	61314&	56500 \\
std dev &39812.026 &	15701.39 &	12972.443\\
\# deadlocks& 7 & 9 & 11\\
			\hline 
		\end{tabular}}
	
\end{table}

  \begin{table}
    \centering
    \caption{ {Wins, draws and defeats on a new railway with 155 resources}}
    \begin{tabular}{cccc}
        
    \hline
        \textbf{Win/Draw/Loss} &\textbf{Centralized}&\textbf{Decentralized}&\textbf{Q-Learning}\\\hline
        Centralized & -&32-11-7&47-0-3 \\
        Decentralized &7-11-32&-&47-0-3\\
        Q-Learning&3-0-47 &3-0-47 &-\\ \hline
    \end{tabular}

    \label{tab:WDL_155}
\end{table}

  \begin{table}
    \centering
    \caption{ {Wins, draws and defeats on new railway with $180$ resources}}
    \begin{tabular}{cccc}
        
    \hline
        \textbf{Win/Draw/Loss} &\textbf{Centralized}&\textbf{Decentralized}&\textbf{Q-Learning}\\\hline
        Centralized & -&29-16-5&39-0-11 \\
        Decentralized &5-16-29&-&41-0-9\\
        Q-Learning&11-0-39 &9-0-41  &-\\ \hline
    \end{tabular}

    \label{tab:WDL_180}
\end{table}

  \begin{table}
    \centering
    \caption{ {Wins, draws and defeats on a new railway with $196$ resources}}
    \begin{tabular}{cccc}
        
    \hline
        \textbf{Win/Draw/Loss} &\textbf{Centralized}&\textbf{Decentralized}&\textbf{Q-Learning}\\\hline
        Centralized & -&30-14-6&44-0-6 \\
        Decentralized &6-14-30&-&46-0-4\\
        Q-Learning&6-0-44 &4-0-46  &-\\ \hline
    \end{tabular}

    \label{tab:WDL_196}
\end{table}
\section{Mixed integer programming comparison}
\label{sec:MIP}

Finally, we decided to compare our results for the Centralized approach with those obtained using a mixed integer program (MIP). The formulation is taken from  \cite{sama2017variable} and it models both the routing and scheduling of trains. However, several of the features that were covered by our Reinforcement Learning model were not easy to extend in a mathematical formulation, therefore we had to neglect some of them. For example, a
crucial aspect such as modelling the train length, and therefore its actual occupation of resources. The MIP model would have suffered computationally with this length adjustment, and it would have not been a fair  comparison. For this reason, solutions of the MIP model with a lower value than RL solutions may potentially be infeasible. We report a brief description of the model adopted in  \cite{sama2017variable} in the Appendix.

We ran the tests on the same machine of the previous experiments, and we report them in Table \ref{tab:Comp_MIP}. To recap, the five different test sets where characterized by:
\begin{itemize}
    \item 100 test instances on 137 resources (100 T 137 R)
    \item 200 test instances on 137 resources (200 T 137 R)
    \item 50 test instances on 155 resources (50 T 155 R)
    \item 50 test instances on 180 resources (50 T 180 R)
    \item 50 test instances on 196 resources (50 T 196 R)
\end{itemize}
We ran each test instance for 10 minutes. 
For each of the five test sets we consider:
\begin{itemize}
    \item the number of instances resolved in 10 minutes, where an instance is considered resolved if it finds an incumbent in the given time limit (Solved MIP).
    \item the number of instances solved both by MIP approach and RL, meaning the algorithm does not end with a deadlock (Both solved).
    \item the average delay of the sets according to MIP approach on the instances in the set \textit{Solved MIP} (Avg delay (MIP)).
    \item the average delay of the sets according to MIP approach on the instances in the set \textit{Both solved} (Avg delay 2 (MIP)).
    \item the average delay of the sets according to RL approach on the instances in the set \textit{Both solved} (Avg delay (RL)).
\end{itemize}
As we can see, increasing the number of resources and/or trains  the number of variables also escalates and therefore the number of solved instances solved reasonably decreases. At the same time, the average delay found in the MIP approach becomes higher than the ones found by RL. In fact while the values are lower for the first two test groups, they are significantly higher in the remaining sets.
 {Particularly, for test sets 100 T 137 R and 200 T 137 R that have the less number of resources and trains MIP the average delay obtained by MIP approach is a little less than the Centralized approach. However when the number of resources increases (from 137 to 155, 180 and 196) and even the maximum number of trains is higher (from 10 and 12 to 13 and 15) MIP for test instances 50 T 185 R finds an average delay that is 150\% higher than the average delay found by RL approach while for test sets 50 T 185 R and 50 T 197R the average delay found is twice higher than average delay returned by RL.}
 \begin{table}
    \small
    \caption{ {Wins, draws and defeats on a new railway with $196$ resources}}
    \begin{tabular}{cccccc}
        
    \hline
        Test &Solved (MIP)& Avg delay (MIP) & Avg delay 2 (MIP) & Avg delay (RL) & Both solved\\\hline
        100 T 137 R & 98&8779&8716&9387&93 \\
        200 T 137 R &187&11712&11503&12893&173\\
        50 T 150 R&37&23550&22960&14666&28\\ 
        50 T 185 R&41&30339&27494&16936&28\\
        50 T 197 R& 43&32926 &32504&16379 &33\\
        \hline
    \end{tabular}

    \label{tab:Comp_MIP}
\end{table}

\section{Conclusions}
This study compares the use of Deep Q-learning with linear Q-learning for tackling the train dispatching problem. Two Deep Q-learning approaches were proposed: Decentralized and Centralized. The former considers a train as an agent with a limited perception of the rail network. The latter observes the entire network and  uses a GNN to estimate the rewards, allowing to change the size of railway network without training a new neural network every time. Computational results inspired on data provided by a U.S. class 1 railroad show that the deep approaches perform better than the 
linear case. The generalization to larger problems, both in terms of number of trains and in terms of railway size,  shows room for improvement, both in terms of search strategy and complexity of the tested network and instances. When the instance is generated by the same distribution of the training set, the algorithms prove to deal efficiently with the problem providing solutions in a very short time. This aspect is crucial in train dispatching, as indeed in any online/real-time planning problem, and in our opinion is one of the reasons that makes this research direction interesting. While solution algorithms based on different paradigms (e.g. Optimization, CP) have proven to tackle the problem effectively in certain cases, scaling and computational burden issues are always behind the corner. As shown in Section \ref{sec:MIP}, we compared our Centralized approach with a MILP formulation, showing the discrepancy between the two methods grows with the complexity of the problem.i
 {In future research, we may compare ourselves or include our heuristics into more complex models and algorithms like  \cite{pellegrini2015recife} and   \cite{corman2017integrating}.}
Moreover, Deep Reinforcement Learning (as in general ML-based approaches) has the advantage of shifting the computational burden to the algorithm's learning stage. Unlike in enumerative algorithms, online response time then becomes basically negligible. In other words, under the assumption that implicitly mapping space-actions-rewards (which presents its own, clear scaling issues) can be done effectively, Deep RL could represent a breakthrough in this application. For this reason, and others listed in the Contribution paragraph in Section \ref{sec:intro}, we believe that it is worth investigating and further advancing this research direction.

\bibliographystyle{plain}
\bibliography{main.bib}




\section*{Appendix}

We represent the railway network as a graph $G=(N,F,A)$ where each node $n \in N$ is a resource. Moreover we have two types of arcs: fixed and disjunctive. For each train and each routing, fixed arcs connects two subsequent resources while disjunctive arcs represent logic alternatives where at most one of them can be activated. A disjunctive arc represents precedence constraints on resources that can be shared by two trains.

The first group of variables is represented by the
routing variables $y_{um}$,  that are assigned to   one if routing $m$ for train $u$ is chosen, and are zero otherwise. The second group are 
disjunctive variables $x_{(krj,ump),(umi,krp)}$, that are driven to one if train $k$ on routing $r$ enters in resource $p$ before train $u$ that chooses routing $m$, and zero otherwise. Finally, the last group of variables $t_{krj}$ are associated with time, in particular the instant when train $k$ enters in resource $j$ following routing $r$.

The resulting mathematical formulation is as follows:

     $$ 
     \left\{
     \begin{array}{rlr}
        \min  &  \displaystyle  \sum_{n \in T}t_{n}p_{n}  \\ \\
         \text{s.t.} & t_{krj}-t_{krp}\geq w_{krp,krj}+M(1-y_{kr}), \hfill \forall\ (krp,krj)\in F  &(i)  \\ \\
         & t_{ump}-t_{krj}\geq w_{krj,ump}^A+M(2-y_{um}-y_{kr}) +Mx_{(krj,ump),(umi,krp)}, \\  &\hfill\forall\   ((krj,ump),(umi,krp))\in A  &(ii) \\ \\
         & t_{krp}-t_{umi}\geq w_{krj,krp}+M(2-y_{um}-y_{kr})+M(1-x_{(krj,ump),(umi,krp)}), \\ &  \hfill\forall\  ((krj,ump),(umi,krp))\in A & (iii)  \\ \\
         & \displaystyle \sum_{a=1}^{R_b} y_{ab} = 1, \hfill \forall\ b\in\left\{1, \dots, Z\right\} & (iv)\\ \\
         & x_{(krj,ump),(umi,krp)} \in \{0,1\},y_{um} \in \{0,1\}, t_k \in \mathbb{R}^+
     \end{array}
     \right.
     $$

Constraints (i) ensure that if train $k$ chooses route $r$ for each resource, then the starting time of two subsequent resources ($i$ and $i+1$) must be at least equal to the time to travel the $i$-th resource. Constraints (ii) -(iii) are precedence constraints for trains that share the same resources.  Group (iv) ensures  each train uses just one routing.

\end{document}

%% file: RL_scheme.tex
\begin{tikzpicture}[->,>=stealth',auto,node distance=1.5cm,
  thick,main node/.style={rectangle, draw,  
    text width=6em, text centered, rounded corners, minimum height=4em}]
    \footnotesize
\node[main node] (1) {Agent};
\node (2) [below of=1] {};
\node[main node] (3) [below of=2] {Environment};
\node[rectangle, draw,  
    text width=2em,  minimum height=0.001em](4) [left of=2,node distance = 3cm ] {};

\draw [->] (1.east) to [out=360,in=360, distance=2.5cm] node {$a_t$} (3.east);
\draw [->] ($(3.west)+( 0, -2mm)$) to [out=180,in=-90] node[near start,below] {$r_{t+1}$}($(4.south)+(-2mm, 0)$);
\draw [->] ($(3.west)+( 0, 2mm)$) to [out=180,in=-90] node[near start,above] {$s_{t+1}$}($(4.south)+(2mm, 0)$);

\draw [->] (4) to [out=90,in=180] node {$s_t$} (1);
\end{tikzpicture}

%% file: grafo.tex
\begin{tikzpicture}
 \node[draw,circle, blue] (a) {1};
 \node [above =0.1cm of a] (tt1) {$m_1 = x_2$};
 \node [above =0.1cm of tt1] (t1) {$x_1 = [0,1,0,1]$};
  \node [below of=a] (d) {};
 \node [below left of=d] (d1) {};
 \node [below right of=d] (d2) {};

  \node[draw,circle, below left of=d1, green] (b) {2};
  \node [below =0.1cm of b] (t2) {$x_2 = [1,0,0,0]$};
   \node [below =0.1cm of t2] (tt2) {$m_2 = x_3$};
   
  \node[draw,circle, below right of=d2, red] (c) {3};
   \node [below =0.1cm of c] (t3) {$x_3 = [0,0,0,0]$};
   \node [below =0.1cm of t3] (tt3) {$m_3 = \frac{x_1 + x_2}{2}$};
  
  \draw [-stealth, green](b) -- (a);
  \draw [-stealth, green](b) -- (c);
  \draw [-stealth, blue](a) -- (c);
  \draw [-stealth, red](c) to [out=150, in=30] (b);
  
\end{tikzpicture}

%% file: network_def.tex
\resizebox{1\textwidth}{!}{%
\begin{tikzpicture}
 \node[draw,rectangle] (x1) {$x_1$};
 \node[draw,rectangle, below =0.75cm  of  x1] (x2) {$x_2$};
 \node[draw,rectangle, below =0.75cm of  x2] (x3) {$x_3$};

  \node[draw,signal, right =0.65cm of x1] (m1) {$M$};
 \node[draw,signal, right =0.65cm of  x2] (m2) {$M$};
 \node[draw,signal, right =0.65cm of  x3] (m3) {$M$};

  \draw [-stealth, green](x2.east) -- (m1.west);
  \draw [-stealth, green](x2.east) -- (m3.west);
  \draw [-stealth, blue](x1.east) -- (m3.west);
  \draw [-stealth, red](x3.east) -- (m2.west);
 
   \node[draw,diamond, right =0.25cm of m1] (mm1) {$m_1$};
 \node[draw,diamond, right =0.25cm of  m2] (mm2) {$m_2$};
 \node[draw,diamond, right =0.25cm of  m3] (mm3) {$m_3$};
 
  \draw [-stealth, green](m2) -- (mm2);
  \draw [-stealth, blue](m1) -- (mm1);
  \draw [-stealth, red](m3) -- (mm3);

    \node[draw,rectangle, right =0.25cm of mm1] (xm1) {$(x_1, m_1)$};
 \node[draw,rectangle, right =0.25cm of  mm2] (xm2) {$(x_2, m_2)$};
 \node[draw,rectangle, right =0.25cm of  mm3] (xm3) {$(x_3, m_3)$};
 
  \draw [-stealth, green](mm2) -- (xm2);
  \draw [-stealth, blue](mm1) -- (xm1);
  \draw [-stealth, red](mm3) -- (xm3);

   \node[draw,signal, right = 0.25cm of xm1] (r1) {$ReLU_1$};
 \node[draw,signal, right =0.25cm of  xm2] (r2) {$ReLU_1$};
 \node[draw,signal, right =0.25cm of  xm3] (r3) {$ReLU_1$};
 
 \draw [-stealth, green](xm2) -- (r2);
  \draw [-stealth, blue](xm1) -- (r1);
  \draw [-stealth, red](xm3) -- (r3);
 
 \node[draw,diamond, right = 0.25cm of r1] (l1) {$l_1$};
 \node[draw,diamond, right =0.25cm of  r2] (l2) {$l_2$};
 \node[draw,diamond, right =0.25cm of  r3] (l3) {$l_3$};
 
  \draw [-stealth, green](r2) -- (l2);
  \draw [-stealth, blue](r1) -- (l1);
  \draw [-stealth, red](r3) -- (l3);

 \node[draw,circle, right =0.65cm of l1] (o1) {$+$};
 \node[draw,circle, right =0.65cm of  l2] (o2) {$+$};
 \node[draw,circle, right =0.65cm of  l3] (o3) {$+$};

 \draw [-stealth, green](l2.east) -- (o1.west);
  \draw [-stealth, green](l2.east) -- (o3.west);
  \draw [-stealth, blue](l1.east) -- (o3.west);
  \draw [-stealth, red](l3.east) -- (o2.west);
  
   \draw [-stealth, green](l2) -- (o2);
  \draw [-stealth, blue](l1) -- (o1);
  \draw [-stealth, red](l3) -- (o3);
 
  \node[draw,signal, right =0.25cm of o1] (rr1) {$ReLU_2$};
 \node[draw,signal, right =0.25cm of  o2] (rr2) {$ReLU_2$};
 \node[draw,signal, right =0.25cm of  o3] (rr3) {$ReLU_2$};

    \draw [-stealth, green](o2) -- (rr2);
  \draw [-stealth, blue](o1) -- (rr1);
  \draw [-stealth, red](o3) -- (rr3);
 
 \node[draw,diamond, right =0.25cm of rr1] (h1) {$h_1$};
 \node[draw,diamond, right =0.25cm of rr2] (h2) {$h_2$};
 \node[draw,diamond, right =0.25cm of rr3] (h3) {$h_3$};

     \draw [-stealth, green](rr2) -- (h2);
  \draw [-stealth, blue](rr1) -- (h1);
  \draw [-stealth, red](rr3) -- (h3);
 
 \node[draw,circle, right =0.25cm of h2] (oo) {$+$};
 
     \draw [-stealth, green](h2) -- (oo);
  \draw [-stealth, blue](h1) -- (oo);
  \draw [-stealth, red](h3) -- (oo);
 
  \node[draw,signal, right =0.25cm of oo] (ooo) {$f$};
    \draw [-stealth](oo) -- (ooo);
   \node[draw,diamond, right =0.25cm of ooo] (oooo) {$o$};
 \draw [-stealth](ooo) -- (oooo);


   
  
  
\end{tikzpicture}
}%

%% file: network_model.tex
\def\layersep{0.8cm}

\begin{tikzpicture}[->,>=stealth',auto,node distance=1.5cm,start chain=going below,
  thick,main node/.style={rectangle,  
    text width=6em, text centered, rounded corners},
    punktchain/.style = {rectangle, rounded corners, draw, very thick,
                     text width=3.1cm, minimum height=3em, inner sep=2pt,
                     align=center, on chain},
   inclass/.style = {punktchain, minimum height=6mm,
                     top color=red!15, bottom color=red!5}]
  \footnotesize    
    
\node[main node] (1) {input};
\node (2) [below of=1] {fully connected };
\node (3) [below of=2,node distance=\layersep] {ReLU};
\node (4) [below of=3] {fully connected };
\node (5) [below of=4,node distance=\layersep] {ReLU};
\node[diamond,draw,  thin,
         inner ysep=0mm, inner xsep=0mm,
         minimum width=3cm] (8) [below of=5] {multiply};
\node[main node] (9) [below of=8] {output};
\node[draw,  thin, rounded corners,
         inner ysep=0mm, inner xsep=0mm,
         minimum width=3cm, minimum height = 1.375cm] (10) at (4cm,-3.18cm) {action mask};

\node   [draw,  thick, rounded corners,
         inner ysep=2.5mm, inner xsep=4mm,
         minimum width=3.2cm, fit= (2) (5.south -| 2.west)  (5.south -| 2.east) (2)] {};
         \node   [draw,  thin, rounded corners,
         inner ysep=0mm, inner xsep=0mm,
         minimum width=3cm, fit= (2) (3.south -| 2.west)  (3.south -| 2.east) (2)] {};
         \node   [draw,  thin, rounded corners,
         inner ysep=0mm, inner xsep=0mm,
         minimum width=3cm, fit= (4) (5.south -| 4.west)  (5.south -| 4.east) (4)] {};

\draw [->] (1) -- (2);
\draw [->] (2) -- (3);
\draw [->] (3) -- (4);
\draw [->] (4) -- (5);
\draw [->] (5) -- (8);
\draw [->] (8) -- (9);

\draw [->] (1.east) to [out=0, in = 90] (10.north);
\draw [->] (10.south) to [out=-90, in = 0] (8.east);


\end{tikzpicture}

%% file: SheetV1.tex
\begin{tikzpicture}

\definecolor{darkgray176}{RGB}{176,176,176}
\definecolor{darkorange25512714}{RGB}{255,127,14}
\definecolor{forestgreen4416044}{RGB}{44,160,44}
\definecolor{lightgray204}{RGB}{204,204,204}
\definecolor{steelblue31119180}{RGB}{31,119,180}

\begin{axis}[
legend cell align={left},
legend style={
  fill opacity=0.8,
  draw opacity=1,
  text opacity=1,
  at={(0.97,0.03)},
  anchor=south east,
  draw=lightgray204
},
log basis x={10},
tick align=outside,
tick pos=left,
x grid style={darkgray176},
xmin=0.696965965672849, xmax=1961.35861337261,
xmode=log,
xtick style={color=black},
y grid style={darkgray176},
ymin=-0.05, ymax=1.05,
ytick style={color=black}
]
\addplot [semithick, steelblue31119180]
table {%
1 0
1 0
1 0
1 0
1 0
1 0
1 0
1 0
1 0
1 0
1 0
1 0
1 0
1 0
1 0
1 0
1 0
1 0
1 0
1 0
1 0
1 0
1 0
1 0
1 0
1 0
1 0
1 0
1 0
1 0
1 0
1 0
1 0
1 0
1 0
1 0
1 0
1 0
1 0
1 0
1 0
1 0
1 0
1 0
1 0
1 0
1 0
1 0
1 0
1 0
1 0
1 0
1 0
1 0
1 0
1 0
1 0
1 0
1 0
1 0
1 0
1 0
1 0
1 0
1 0
1 0
1 0
1 0
1 0
1 0
1 0
1 0
1 0
1 0
1 0
1 0
1 0
1 0
1 0
1 0
1 0
1 0
1 0
1 0
1 0
1 0
1 0
1 0
1 0
1 0
1 0
1 0
1 0
1 0
1 0
1 0
1 0
1 0
1 0
1 0
1 0
1 0
1 0
1 0
1 0
1 0
1 0
1.00030611644761 0.05
1.00795046685772 0.05
1.0139944022391 0.05
1.01556617184563 0.05
1.01637712202916 0.05
1.02540358928132 0.05
1.03121452894438 0.05
1.03884681269306 0.05
1.04234851914618 0.05
1.04412416851441 0.05
1.07342401191546 0.05
1.08119112990797 0.05
1.0829709554412 0.05
1.090128056899 0.05
1.09855486434761 0.05
1.0989103101425 0.05
1.11860292988741 0.05
1.11937748505819 0.05
1.12440159108088 0.05
1.12464287946854 0.05
1.13761428306915 0.05
1.14453877706929 0.05
1.1486301369863 0.05
1.15170670037927 0.05
1.15432058797248 0.05
1.15687898616748 0.05
1.1642694483424 0.05
1.16970333159813 0.05
1.16987570781427 0.05
1.19160619729359 0.05
1.1935744138695 0.05
1.19916153181938 0.05
1.20545378698512 0.05
1.22292768959436 0.05
1.22951837349398 0.05
1.23601220752798 0.05
1.24123250108241 0.05
1.24712893316729 0.05
1.24832887615409 0.05
1.24928425357873 0.05
1.26330613911574 0.05
1.26532864989015 0.05
1.29723400074631 0.06
1.29781858994625 0.06
1.30225713854817 0.06
1.32113741705711 0.06
1.33371577176075 0.06
1.35630848326089 0.06
1.35727882542635 0.06
1.36753753398747 0.06
1.36915016665221 0.06
1.37461323013416 0.06
1.39145029337804 0.07
1.39701954320124 0.08
1.42563331967549 0.08
1.43330363663667 0.09
1.43938575875824 0.09
1.45330535152151 0.09
1.45380125179153 0.09
1.46672428694901 0.09
1.47739373702734 0.09
1.47903935468101 0.09
1.49678373911947 0.09
1.51654846335697 0.09
1.56316117370243 0.09
1.59943449575872 0.09
1.60079435120014 0.09
1.60362208939245 0.09
1.61730249324121 0.09
1.61850649350649 0.09
1.69188854805726 0.09
1.74198427102238 0.1
1.77451210909946 0.1
1.78895802025482 0.1
1.78916037155844 0.1
1.79047760369366 0.1
1.85136166963604 0.11
1.86009969433341 0.11
1.91884403226908 0.12
1.93623543838136 0.12
1.9536436849342 0.12
1.97294250281849 0.12
2.00782640767303 0.12
2.04053257713249 0.13
2.10310285460993 0.14
2.12620103473762 0.14
2.28794500577208 0.14
2.29955681448633 0.14
2.32421107017544 0.15
2.33437120143266 0.15
2.39144402239104 0.16
2.42308961937716 0.17
2.46103613605954 0.18
2.46408581827749 0.19
2.46823636560032 0.2
2.51363636363636 0.21
2.53574604303243 0.21
2.60459954918434 0.22
2.61774484334054 0.22
2.67945997422732 0.23
2.68678880130629 0.24
2.69754427458804 0.25
2.784387739695 0.25
2.79734457793234 0.25
2.85550219459267 0.26
2.89473663490014 0.27
2.90161965739961 0.28
2.92761170283373 0.29
2.92761170283373 0.29
3.003608630394 0.29
3.0593667221085 0.3
3.15342759806988 0.31
3.20106666624543 0.32
3.20738701298701 0.33
3.33910034602076 0.34
3.49696373416623 0.34
3.56094182527486 0.35
3.6773054964539 0.36
3.73212047396444 0.37
3.74408117249154 0.38
3.83642674628708 0.38
4.08141256079509 0.39
4.11983958780548 0.39
4.3204483214649 0.4
4.33470387471845 0.41
4.4291651687421 0.42
4.50574254877543 0.43
4.50733416646297 0.44
4.56597639019793 0.45
4.73754351483776 0.46
4.95142989343536 0.47
5.05889242478141 0.48
5.11000186457311 0.49
5.1315963060686 0.5
5.1768799237214 0.5
5.29402185560274 0.51
5.41292492144229 0.52
5.44682005089058 0.53
5.54233114651671 0.53
5.55354763135228 0.53
5.600793601359 0.54
5.99549459297086 0.55
6.2276461335648 0.56
6.27817676205951 0.57
6.37217494428526 0.58
6.4523272985244 0.59
6.47030279255319 0.6
6.5039863274065 0.61
6.50958433448984 0.62
6.76083227377596 0.63
7.01013925654745 0.64
7.07675127615843 0.65
7.32750939973615 0.65
7.46580280161535 0.66
7.4755049583472 0.67
7.51810108964078 0.68
7.55878836915297 0.69
7.66727185638359 0.7
7.68061177834838 0.71
8.03308938661237 0.72
8.13676514742397 0.73
8.15788055315471 0.74
8.72267895496796 0.75
8.85276196319018 0.76
9.01045146155973 0.76
9.02088179970436 0.76
9.2294249066981 0.77
9.35376881849315 0.78
9.37822989581761 0.79
9.9573826458037 0.8
10.1073559916214 0.81
10.2867434842942 0.82
10.3431845454545 0.83
10.353088042328 0.84
10.6010276670164 0.85
10.6077700137282 0.86
10.9416656246092 0.87
11.1218375507654 0.87
11.337235658559 0.88
11.3768922654125 0.89
11.3834379754601 0.9
13.487374754386 0.91
13.8911720994466 0.92
16.0501178684711 0.93
16.7033498768092 0.94
22.0849318525997 0.94
32.0188315521629 0.94
36.0449852364475 0.95
42.8466300613497 0.96
94.3275862068966 0.97
105.422289156627 0.97
515.362712068965 0.98
1367.0002 0.99
};
\addlegendentry{Q-learning}
\addplot [semithick, darkorange25512714]
table {%
1 0
1 0
1 0
1 0
1 0
1 0
1 0
1 0
1 0
1 0
1 0
1 0
1 0
1 0
1 0
1 0
1 0
1 0
1 0
1 0
1 0
1 0
1 0
1 0
1 0
1 0
1 0
1 0
1 0
1 0
1 0
1 0
1 0
1 0
1 0
1 0
1 0
1 0
1 0
1 0
1 0
1 0
1 0
1 0
1 0
1 0
1 0
1 0
1 0
1 0
1 0
1 0
1 0
1 0
1 0
1 0
1 0
1 0
1 0
1 0
1 0
1 0
1 0
1 0
1 0
1 0
1 0
1 0
1 0
1 0
1 0
1 0
1 0
1 0
1 0
1 0
1 0
1 0
1 0
1 0
1 0
1 0
1 0
1 0
1 0
1 0
1 0
1 0
1 0
1 0
1 0
1 0
1 0
1 0
1 0
1 0
1 0
1 0
1 0
1 0
1 0
1 0
1 0
1 0
1 0
1 0
1 0
1.00030611644761 0.42
1.00795046685772 0.42
1.0139944022391 0.43
1.01556617184563 0.44
1.01637712202916 0.44
1.02540358928132 0.45
1.03121452894438 0.45
1.03884681269306 0.45
1.04234851914618 0.46
1.04412416851441 0.47
1.07342401191546 0.48
1.08119112990797 0.49
1.0829709554412 0.49
1.090128056899 0.5
1.09855486434761 0.51
1.0989103101425 0.51
1.11860292988741 0.51
1.11937748505819 0.52
1.12440159108088 0.53
1.12464287946854 0.53
1.13761428306915 0.54
1.14453877706929 0.54
1.1486301369863 0.54
1.15170670037927 0.55
1.15432058797248 0.56
1.15687898616748 0.57
1.1642694483424 0.58
1.16970333159813 0.59
1.16987570781427 0.6
1.19160619729359 0.61
1.1935744138695 0.62
1.19916153181938 0.62
1.20545378698512 0.63
1.22292768959436 0.64
1.22951837349398 0.65
1.23601220752798 0.66
1.24123250108241 0.66
1.24712893316729 0.67
1.24832887615409 0.68
1.24928425357873 0.69
1.26330613911574 0.69
1.26532864989015 0.7
1.29723400074631 0.7
1.29781858994625 0.7
1.30225713854817 0.71
1.32113741705711 0.71
1.33371577176075 0.72
1.35630848326089 0.72
1.35727882542635 0.73
1.36753753398747 0.73
1.36915016665221 0.73
1.37461323013416 0.73
1.39145029337804 0.73
1.39701954320124 0.73
1.42563331967549 0.73
1.43330363663667 0.73
1.43938575875824 0.74
1.45330535152151 0.74
1.45380125179153 0.74
1.46672428694901 0.75
1.47739373702734 0.75
1.47903935468101 0.75
1.49678373911947 0.75
1.51654846335697 0.76
1.56316117370243 0.77
1.59943449575872 0.78
1.60079435120014 0.78
1.60362208939245 0.79
1.61730249324121 0.8
1.61850649350649 0.8
1.69188854805726 0.8
1.74198427102238 0.8
1.77451210909946 0.81
1.78895802025482 0.81
1.78916037155844 0.82
1.79047760369366 0.82
1.85136166963604 0.82
1.86009969433341 0.82
1.91884403226908 0.82
1.93623543838136 0.83
1.9536436849342 0.84
1.97294250281849 0.84
2.00782640767303 0.84
2.04053257713249 0.84
2.10310285460993 0.84
2.12620103473762 0.85
2.28794500577208 0.86
2.29955681448633 0.86
2.32421107017544 0.86
2.33437120143266 0.87
2.39144402239104 0.87
2.42308961937716 0.87
2.46103613605954 0.87
2.46408581827749 0.87
2.46823636560032 0.87
2.51363636363636 0.87
2.53574604303243 0.88
2.60459954918434 0.88
2.61774484334054 0.89
2.67945997422732 0.89
2.68678880130629 0.89
2.69754427458804 0.89
2.784387739695 0.89
2.79734457793234 0.89
2.85550219459267 0.89
2.89473663490014 0.89
2.90161965739961 0.89
2.92761170283373 0.89
2.92761170283373 0.89
3.003608630394 0.9
3.0593667221085 0.9
3.15342759806988 0.9
3.20106666624543 0.9
3.20738701298701 0.9
3.33910034602076 0.9
3.49696373416623 0.91
3.56094182527486 0.91
3.6773054964539 0.91
3.73212047396444 0.91
3.74408117249154 0.91
3.83642674628708 0.92
4.08141256079509 0.92
4.11983958780548 0.93
4.3204483214649 0.93
4.33470387471845 0.93
4.4291651687421 0.93
4.50574254877543 0.93
4.50733416646297 0.93
4.56597639019793 0.93
4.73754351483776 0.93
4.95142989343536 0.93
5.05889242478141 0.93
5.11000186457311 0.93
5.1315963060686 0.93
5.1768799237214 0.94
5.29402185560274 0.94
5.41292492144229 0.94
5.44682005089058 0.94
5.54233114651671 0.95
5.55354763135228 0.96
5.600793601359 0.96
5.99549459297086 0.96
6.2276461335648 0.96
6.27817676205951 0.96
6.37217494428526 0.96
6.4523272985244 0.96
6.47030279255319 0.96
6.5039863274065 0.96
6.50958433448984 0.96
6.76083227377596 0.96
7.01013925654745 0.96
7.07675127615843 0.96
7.32750939973615 0.96
7.46580280161535 0.96
7.4755049583472 0.96
7.51810108964078 0.96
7.55878836915297 0.96
7.66727185638359 0.96
7.68061177834838 0.96
8.03308938661237 0.96
8.13676514742397 0.96
8.15788055315471 0.96
8.72267895496796 0.96
8.85276196319018 0.96
9.01045146155973 0.97
9.02088179970436 0.97
9.2294249066981 0.97
9.35376881849315 0.97
9.37822989581761 0.97
9.9573826458037 0.97
10.1073559916214 0.97
10.2867434842942 0.97
10.3431845454545 0.97
10.353088042328 0.97
10.6010276670164 0.97
10.6077700137282 0.97
10.9416656246092 0.97
11.1218375507654 0.98
11.337235658559 0.98
11.3768922654125 0.98
11.3834379754601 0.98
13.487374754386 0.98
13.8911720994466 0.98
16.0501178684711 0.98
16.7033498768092 0.98
22.0849318525997 0.99
32.0188315521629 1
36.0449852364475 1
42.8466300613497 1
94.3275862068966 1
105.422289156627 1
515.362712068965 1
1367.0002 1
};
\addlegendentry{Decentralized}
\addplot [semithick, forestgreen4416044]
table {%
1 0
1 0
1 0
1 0
1 0
1 0
1 0
1 0
1 0
1 0
1 0
1 0
1 0
1 0
1 0
1 0
1 0
1 0
1 0
1 0
1 0
1 0
1 0
1 0
1 0
1 0
1 0
1 0
1 0
1 0
1 0
1 0
1 0
1 0
1 0
1 0
1 0
1 0
1 0
1 0
1 0
1 0
1 0
1 0
1 0
1 0
1 0
1 0
1 0
1 0
1 0
1 0
1 0
1 0
1 0
1 0
1 0
1 0
1 0
1 0
1 0
1 0
1 0
1 0
1 0
1 0
1 0
1 0
1 0
1 0
1 0
1 0
1 0
1 0
1 0
1 0
1 0
1 0
1 0
1 0
1 0
1 0
1 0
1 0
1 0
1 0
1 0
1 0
1 0
1 0
1 0
1 0
1 0
1 0
1 0
1 0
1 0
1 0
1 0
1 0
1 0
1 0
1 0
1 0
1 0
1 0
1 0
1.00030611644761 0.6
1.00795046685772 0.61
1.0139944022391 0.61
1.01556617184563 0.61
1.01637712202916 0.62
1.02540358928132 0.62
1.03121452894438 0.63
1.03884681269306 0.64
1.04234851914618 0.64
1.04412416851441 0.64
1.07342401191546 0.64
1.08119112990797 0.64
1.0829709554412 0.65
1.090128056899 0.65
1.09855486434761 0.65
1.0989103101425 0.66
1.11860292988741 0.67
1.11937748505819 0.67
1.12440159108088 0.67
1.12464287946854 0.68
1.13761428306915 0.68
1.14453877706929 0.69
1.1486301369863 0.7
1.15170670037927 0.7
1.15432058797248 0.7
1.15687898616748 0.7
1.1642694483424 0.7
1.16970333159813 0.7
1.16987570781427 0.7
1.19160619729359 0.7
1.1935744138695 0.7
1.19916153181938 0.71
1.20545378698512 0.71
1.22292768959436 0.71
1.22951837349398 0.71
1.23601220752798 0.71
1.24123250108241 0.72
1.24712893316729 0.72
1.24832887615409 0.72
1.24928425357873 0.72
1.26330613911574 0.73
1.26532864989015 0.73
1.29723400074631 0.73
1.29781858994625 0.74
1.30225713854817 0.74
1.32113741705711 0.75
1.33371577176075 0.75
1.35630848326089 0.76
1.35727882542635 0.76
1.36753753398747 0.77
1.36915016665221 0.78
1.37461323013416 0.79
1.39145029337804 0.79
1.39701954320124 0.79
1.42563331967549 0.8
1.43330363663667 0.8
1.43938575875824 0.8
1.45330535152151 0.81
1.45380125179153 0.82
1.46672428694901 0.82
1.47739373702734 0.83
1.47903935468101 0.84
1.49678373911947 0.85
1.51654846335697 0.85
1.56316117370243 0.85
1.59943449575872 0.85
1.60079435120014 0.86
1.60362208939245 0.86
1.61730249324121 0.86
1.61850649350649 0.87
1.69188854805726 0.88
1.74198427102238 0.88
1.77451210909946 0.88
1.78895802025482 0.89
1.78916037155844 0.89
1.79047760369366 0.9
1.85136166963604 0.9
1.86009969433341 0.91
1.91884403226908 0.91
1.93623543838136 0.91
1.9536436849342 0.91
1.97294250281849 0.92
2.00782640767303 0.93
2.04053257713249 0.93
2.10310285460993 0.93
2.12620103473762 0.93
2.28794500577208 0.93
2.29955681448633 0.94
2.32421107017544 0.94
2.33437120143266 0.94
2.39144402239104 0.94
2.42308961937716 0.94
2.46103613605954 0.94
2.46408581827749 0.94
2.46823636560032 0.94
2.51363636363636 0.94
2.53574604303243 0.94
2.60459954918434 0.94
2.61774484334054 0.94
2.67945997422732 0.94
2.68678880130629 0.94
2.69754427458804 0.94
2.784387739695 0.95
2.79734457793234 0.96
2.85550219459267 0.96
2.89473663490014 0.96
2.90161965739961 0.96
2.92761170283373 0.96
2.92761170283373 0.96
3.003608630394 0.97
3.0593667221085 0.97
3.15342759806988 0.97
3.20106666624543 0.97
3.20738701298701 0.97
3.33910034602076 0.97
3.49696373416623 0.97
3.56094182527486 0.97
3.6773054964539 0.97
3.73212047396444 0.97
3.74408117249154 0.97
3.83642674628708 0.97
4.08141256079509 0.97
4.11983958780548 0.97
4.3204483214649 0.97
4.33470387471845 0.97
4.4291651687421 0.97
4.50574254877543 0.97
4.50733416646297 0.97
4.56597639019793 0.97
4.73754351483776 0.97
4.95142989343536 0.97
5.05889242478141 0.97
5.11000186457311 0.97
5.1315963060686 0.97
5.1768799237214 0.97
5.29402185560274 0.97
5.41292492144229 0.97
5.44682005089058 0.97
5.54233114651671 0.97
5.55354763135228 0.97
5.600793601359 0.97
5.99549459297086 0.97
6.2276461335648 0.97
6.27817676205951 0.97
6.37217494428526 0.97
6.4523272985244 0.97
6.47030279255319 0.97
6.5039863274065 0.97
6.50958433448984 0.97
6.76083227377596 0.97
7.01013925654745 0.97
7.07675127615843 0.97
7.32750939973615 0.98
7.46580280161535 0.98
7.4755049583472 0.98
7.51810108964078 0.98
7.55878836915297 0.98
7.66727185638359 0.98
7.68061177834838 0.98
8.03308938661237 0.98
8.13676514742397 0.98
8.15788055315471 0.98
8.72267895496796 0.98
8.85276196319018 0.98
9.01045146155973 0.98
9.02088179970436 0.99
9.2294249066981 0.99
9.35376881849315 0.99
9.37822989581761 0.99
9.9573826458037 0.99
10.1073559916214 0.99
10.2867434842942 0.99
10.3431845454545 0.99
10.353088042328 0.99
10.6010276670164 0.99
10.6077700137282 0.99
10.9416656246092 0.99
11.1218375507654 0.99
11.337235658559 0.99
11.3768922654125 0.99
11.3834379754601 0.99
13.487374754386 0.99
13.8911720994466 0.99
16.0501178684711 0.99
16.7033498768092 0.99
22.0849318525997 0.99
32.0188315521629 0.99
36.0449852364475 0.99
42.8466300613497 0.99
94.3275862068966 0.99
105.422289156627 1
515.362712068965 1
1367.0002 1
};
\addlegendentry{Centralized}
\end{axis}

\end{tikzpicture}

%% file: SheetV2.tex
\begin{tikzpicture}

\definecolor{darkgray176}{RGB}{176,176,176}
\definecolor{darkorange25512714}{RGB}{255,127,14}
\definecolor{forestgreen4416044}{RGB}{44,160,44}
\definecolor{lightgray204}{RGB}{204,204,204}
\definecolor{steelblue31119180}{RGB}{31,119,180}

\begin{axis}[
legend cell align={left},
legend style={
  fill opacity=0.8,
  draw opacity=1,
  text opacity=1,
  at={(0.97,0.03)},
  anchor=south east,
  draw=lightgray204
},
log basis x={10},
tick align=outside,
tick pos=left,
x grid style={darkgray176},
xmin=0.602504295924316, xmax=41767.3475031307,
xmode=log,
xtick style={color=black},
y grid style={darkgray176},
ymin=-0.05, ymax=1.05,
ytick style={color=black}
]
\addplot [semithick, steelblue31119180]
table {%
1 0
1 0
1 0
1 0
1 0
1 0
1 0
1 0
1 0
1 0
1 0
1 0
1 0
1 0
1 0
1 0
1 0
1 0
1 0
1 0
1 0
1 0
1 0
1 0
1 0
1 0
1 0
1 0
1 0
1 0
1 0
1 0
1 0
1 0
1 0
1 0
1 0
1 0
1 0
1 0
1 0
1 0
1 0
1 0
1 0
1 0
1 0
1 0
1 0
1 0
1 0
1 0
1 0
1 0
1 0
1 0
1 0
1 0
1 0
1 0
1 0
1 0
1 0
1 0
1 0
1 0
1 0
1 0
1 0
1 0
1 0
1 0
1 0
1 0
1 0
1 0
1 0
1 0
1 0
1 0
1 0
1 0
1 0
1 0
1 0
1 0
1 0
1 0
1 0
1 0
1 0
1 0
1 0
1 0
1 0
1 0
1 0
1 0
1 0
1 0
1 0
1 0
1 0
1 0
1 0
1 0
1 0
1 0
1 0
1 0
1 0
1 0
1 0
1 0
1 0
1 0
1 0
1 0
1 0
1 0
1 0
1 0
1 0
1 0
1 0
1 0
1 0
1 0
1 0
1 0
1 0
1 0
1 0
1 0
1 0
1 0
1 0
1 0
1 0
1 0
1 0
1 0
1 0
1 0
1 0
1 0
1 0
1 0
1 0
1 0
1 0
1 0
1 0
1 0
1 0
1 0
1 0
1 0
1 0
1 0
1 0
1 0
1 0
1 0
1 0
1 0
1 0
1 0
1 0
1 0
1 0
1 0
1 0
1 0
1 0
1 0
1 0
1 0
1 0
1 0
1 0
1 0
1 0
1 0
1 0
1 0
1 0
1 0
1 0
1 0
1 0
1 0
1 0
1 0
1 0
1 0
1 0
1 0
1 0
1 0
1 0
1 0
1 0
1 0
1 0
1 0
1 0
1 0
1 0
1 0
1 0
1 0
1 0
1 0
1 0
1 0
1 0
1 0
1 0
1 0
1 0
1 0
1 0
1 0
1 0
1 0
1 0
1 0
1 0
1 0
1 0
1 0
1 0
1 0
1 0
1 0
1 0
1 0
1.00140789247829 0.08
1.00279460038502 0.08
1.00379162604268 0.08
1.00853807283499 0.08
1.00896905829596 0.08
1.00979586684715 0.08
1.01107601600337 0.08
1.01116389548694 0.08
1.01249132546842 0.08
1.01298945298207 0.08
1.01303538175047 0.08
1.01790419566246 0.08
1.02144485495255 0.08
1.02281144781145 0.08
1.02288985966346 0.08
1.02734103769525 0.08
1.02869130844511 0.08
1.02934169278997 0.08
1.03779958826497 0.08
1.04499430771806 0.08
1.04892086330935 0.08
1.05330862890418 0.08
1.05466834303175 0.08
1.05632658768572 0.08
1.05838521033767 0.08
1.06083011079731 0.08
1.06174603174603 0.08
1.06790926427908 0.08
1.07567357966892 0.08
1.0780092650946 0.08
1.07879253004147 0.08
1.08238609515997 0.08
1.08354837320105 0.08
1.08574135480815 0.08
1.08829570841889 0.08
1.09051165595912 0.08
1.0950463149416 0.08
1.09521045880803 0.08
1.09530042764255 0.08
1.09600008333333 0.08
1.10135514303828 0.08
1.10408921933085 0.08
1.1062818336163 0.08
1.11002444987775 0.08
1.1120019050643 0.08
1.11484593837535 0.08
1.122725871107 0.08
1.1310381294964 0.08
1.13237667401507 0.08
1.13480013550202 0.08
1.14012556903766 0.08
1.1406333870102 0.08
1.14192766759401 0.08
1.14639597690445 0.08
1.15068493150685 0.08
1.15238593745839 0.08
1.15249994299071 0.08
1.16362449139182 0.08
1.17492253066667 0.085
1.18231248538695 0.085
1.19492127342914 0.085
1.19637298805815 0.085
1.19726812371134 0.085
1.20735314612275 0.085
1.21225155665616 0.085
1.2147406733394 0.085
1.21481225524073 0.085
1.21952569460622 0.085
1.22072602459016 0.085
1.23001572645559 0.085
1.23203222174937 0.085
1.24193036418915 0.085
1.24391850355669 0.085
1.24817857345438 0.085
1.25060829396074 0.085
1.25384044644401 0.085
1.25871171754548 0.085
1.26674237879034 0.09
1.26676392362633 0.09
1.27864940493484 0.09
1.28797424051519 0.09
1.2939824537037 0.09
1.29656794907279 0.09
1.29796239126988 0.09
1.3026712823864 0.09
1.3200782680048 0.09
1.33157540394973 0.09
1.33220255231158 0.09
1.33641160949868 0.09
1.33985710290546 0.09
1.33985710290546 0.09
1.34871794871795 0.09
1.3582554517134 0.09
1.37442922374429 0.09
1.3799374169279 0.09
1.38752052545156 0.095
1.40692260044186 0.095
1.4089807836379 0.095
1.41257113295396 0.095
1.41351805205709 0.095
1.42471042471042 0.095
1.44185560968721 0.095
1.44358090769118 0.095
1.47236171656019 0.095
1.48862115127175 0.095
1.49384038564542 0.095
1.50178583766449 0.095
1.52545113547377 0.095
1.53564277711114 0.1
1.53852715382428 0.105
1.54213081369823 0.105
1.55835038463637 0.105
1.55918926465238 0.105
1.56451988330341 0.105
1.56869936305732 0.11
1.57169484171322 0.115
1.57484875757574 0.12
1.5977358490566 0.125
1.60845701715762 0.125
1.6242652558978 0.125
1.63279248505551 0.125
1.64080873950139 0.125
1.64080873950139 0.125
1.64108675721562 0.125
1.64236718124643 0.13
1.67110571390557 0.13
1.67686547895688 0.13
1.68555049261084 0.13
1.69003816004355 0.135
1.71669884169884 0.135
1.72125977191489 0.135
1.72204319459042 0.135
1.72204319459042 0.135
1.73183109348684 0.135
1.73856209150327 0.135
1.73856209150327 0.135
1.73918368677634 0.135
1.75845339422148 0.14
1.77120822622108 0.145
1.79557740835115 0.145
1.81331314821576 0.145
1.8551444152249 0.15
1.86378635862486 0.155
1.89261330525837 0.155
1.91658291457286 0.155
1.92810324315206 0.155
1.93075127435604 0.155
1.93410887289448 0.155
1.93563947751611 0.155
1.98 0.16
1.98569076599327 0.165
1.98716290322581 0.17
2.02559867877787 0.17
2.06079983318953 0.175
2.0616903978688 0.18
2.08114230210123 0.18
2.10430562506503 0.185
2.14128506724684 0.19
2.15144872695347 0.195
2.15962776477199 0.2
2.17139242783505 0.205
2.17407969324928 0.21
2.18582897340024 0.215
2.21248589695374 0.22
2.23233479986583 0.22
2.28119504685994 0.225
2.29514858874071 0.23
2.31623960398644 0.235
2.34471080114449 0.235
2.46124486821645 0.24
2.50234437478026 0.245
2.54240342380899 0.25
2.59858747420225 0.25
2.60248946551091 0.255
2.61857323015873 0.26
2.62017850814518 0.265
2.62958539119804 0.27
2.63205885867512 0.275
2.63340821990221 0.275
2.6553686252648 0.275
2.66458229114557 0.275
2.79926699418766 0.275
2.8071381887933 0.28
2.82906688687036 0.285
2.82966331175309 0.285
2.83612594557693 0.285
2.88133294587092 0.285
2.88201543165468 0.29
2.90796478506769 0.295
2.94802898915482 0.295
2.96410638960314 0.3
2.96837130131681 0.305
2.97033898305085 0.31
2.9846700389233 0.315
3.00147000419815 0.32
3.01240857763267 0.325
3.02698055370123 0.325
3.02796522572436 0.33
3.04781023725036 0.335
3.05076910005294 0.335
3.08855169946333 0.34
3.09554582215065 0.345
3.10440670666189 0.35
3.1658608845888 0.355
3.16732477788746 0.36
3.19796557120501 0.36
3.20588357080967 0.36
3.21909527275224 0.365
3.24057375228896 0.37
3.24900227472596 0.37
3.31825385824507 0.375
3.31930250549153 0.38
3.35261220958972 0.385
3.38899671814672 0.39
3.42154872018385 0.395
3.42460612279802 0.4
3.43574921404464 0.405
3.51051914985342 0.41
3.51402941176471 0.415
3.52601916260679 0.42
3.66475635876841 0.425
3.68221511814957 0.43
3.6825752327914 0.435
3.73810134157132 0.44
3.76100793874391 0.445
3.82801719274097 0.45
3.84675739698557 0.455
3.86223346793349 0.46
3.90545304601627 0.465
3.91859374874372 0.47
3.94198372520865 0.475
3.99458638405707 0.48
3.99716089124534 0.485
4.03676754382979 0.49
4.07066065681614 0.495
4.10749208251372 0.5
4.12235289013895 0.505
4.12453337484433 0.51
4.16345750857381 0.515
4.18424826090482 0.52
4.19038088453621 0.525
4.29361777564626 0.53
4.31226040515654 0.535
4.36831384767498 0.54
4.41424392352105 0.545
4.43021311674716 0.55
4.47712200075216 0.555
4.49304529765886 0.56
4.51111111111111 0.565
4.58961142857143 0.565
4.71868028301887 0.57
4.74943269848734 0.575
4.82877161088405 0.58
4.95103531050955 0.585
5.05653009004502 0.59
5.36861359372813 0.595
5.41371495271868 0.6
5.50454139216618 0.605
5.61927910150697 0.61
5.64347023911187 0.615
5.66520140293961 0.62
5.67138455008489 0.625
5.7172490862423 0.63
5.74192375210692 0.635
5.7944063771518 0.64
5.84623600552868 0.645
5.92644768932018 0.65
6.06668544017296 0.655
6.16879451252895 0.66
6.20837482953296 0.665
6.21431991732711 0.67
6.21431991732711 0.67
6.24325325872284 0.67
6.2557923503861 0.675
6.26902996375352 0.68
6.32113694762684 0.685
6.33118291377718 0.69
6.39088494835219 0.695
6.60209476439791 0.7
6.62922038154248 0.705
6.67407481481481 0.71
6.7505980013035 0.715
6.79605413231066 0.72
6.84771066389964 0.725
6.87053019061051 0.73
6.92431147594671 0.735
6.92431147594671 0.735
7.02771102642041 0.74
7.0703344 0.745
7.07247365369606 0.75
7.24802053835804 0.755
7.24802053835804 0.755
7.29669833267304 0.76
7.34546383326968 0.76
7.50535055749521 0.765
7.52929229188078 0.77
7.95458560411311 0.775
8.07105606258149 0.78
8.15290387743683 0.785
8.23619968331987 0.79
8.26659608801251 0.795
8.52526693447599 0.795
8.57800134728033 0.8
8.95905296957672 0.805
9.06703997837469 0.81
9.138502118613 0.815
9.14117647058823 0.82
9.19497104059463 0.82
9.27184124472136 0.825
9.27184124472136 0.825
9.44323963397198 0.83
9.46920699921789 0.835
9.47410512820513 0.84
9.5886230043499 0.845
9.5899465506836 0.85
9.62998818609514 0.855
9.70272713032778 0.86
10.0047281323877 0.865
11.1268536574074 0.865
11.3147221376039 0.87
11.3147221376039 0.87
12.1358851583113 0.875
12.448430941704 0.88
12.5156926927891 0.885
12.9678954720088 0.885
13.6333176026428 0.89
13.6333176026428 0.89
14.553184921067 0.895
15.1614041461007 0.9
17.1414161392405 0.905
17.3913712374582 0.905
18.5768076593312 0.905
21.0440566082425 0.905
21.2749365845675 0.91
22.0776293759513 0.915
23.2808918338109 0.92
29.5187319646208 0.92
31.2101575999066 0.92
34.750181438687 0.92
38.0563524590164 0.925
42.3937714006196 0.93
43.3539377688299 0.93
49.7418770009789 0.935
50.9882411764706 0.935
52.0018547479126 0.94
52.4724478326985 0.945
117.0001 0.945
234.0001 0.945
329.000100000299 0.95
489.0001 0.95
591.0002 0.95
1020.0001 0.95
1375.0003 0.955
1531.0005 0.96
1936.0005 0.965
2237.0001 0.965
2308.0006 0.97
3073.0012 0.975
4211.0003 0.98
9161.001 0.985
12283.0014000112 0.99
25165.0063 0.995
};
\addlegendentry{Q-learning}
\addplot [semithick, darkorange25512714]
table {%
1 0
1 0
1 0
1 0
1 0
1 0
1 0
1 0
1 0
1 0
1 0
1 0
1 0
1 0
1 0
1 0
1 0
1 0
1 0
1 0
1 0
1 0
1 0
1 0
1 0
1 0
1 0
1 0
1 0
1 0
1 0
1 0
1 0
1 0
1 0
1 0
1 0
1 0
1 0
1 0
1 0
1 0
1 0
1 0
1 0
1 0
1 0
1 0
1 0
1 0
1 0
1 0
1 0
1 0
1 0
1 0
1 0
1 0
1 0
1 0
1 0
1 0
1 0
1 0
1 0
1 0
1 0
1 0
1 0
1 0
1 0
1 0
1 0
1 0
1 0
1 0
1 0
1 0
1 0
1 0
1 0
1 0
1 0
1 0
1 0
1 0
1 0
1 0
1 0
1 0
1 0
1 0
1 0
1 0
1 0
1 0
1 0
1 0
1 0
1 0
1 0
1 0
1 0
1 0
1 0
1 0
1 0
1 0
1 0
1 0
1 0
1 0
1 0
1 0
1 0
1 0
1 0
1 0
1 0
1 0
1 0
1 0
1 0
1 0
1 0
1 0
1 0
1 0
1 0
1 0
1 0
1 0
1 0
1 0
1 0
1 0
1 0
1 0
1 0
1 0
1 0
1 0
1 0
1 0
1 0
1 0
1 0
1 0
1 0
1 0
1 0
1 0
1 0
1 0
1 0
1 0
1 0
1 0
1 0
1 0
1 0
1 0
1 0
1 0
1 0
1 0
1 0
1 0
1 0
1 0
1 0
1 0
1 0
1 0
1 0
1 0
1 0
1 0
1 0
1 0
1 0
1 0
1 0
1 0
1 0
1 0
1 0
1 0
1 0
1 0
1 0
1 0
1 0
1 0
1 0
1 0
1 0
1 0
1 0
1 0
1 0
1 0
1 0
1 0
1 0
1 0
1 0
1 0
1 0
1 0
1 0
1 0
1 0
1 0
1 0
1 0
1 0
1 0
1 0
1 0
1 0
1 0
1 0
1 0
1 0
1 0
1 0
1 0
1 0
1 0
1 0
1 0
1 0
1 0
1 0
1 0
1 0
1 0
1.00140789247829 0.485
1.00279460038502 0.49
1.00379162604268 0.495
1.00853807283499 0.5
1.00896905829596 0.5
1.00979586684715 0.505
1.01107601600337 0.505
1.01116389548694 0.505
1.01249132546842 0.505
1.01298945298207 0.51
1.01303538175047 0.515
1.01790419566246 0.52
1.02144485495255 0.525
1.02281144781145 0.53
1.02288985966346 0.53
1.02734103769525 0.535
1.02869130844511 0.535
1.02934169278997 0.535
1.03779958826497 0.535
1.04499430771806 0.535
1.04892086330935 0.54
1.05330862890418 0.545
1.05466834303175 0.545
1.05632658768572 0.545
1.05838521033767 0.545
1.06083011079731 0.545
1.06174603174603 0.55
1.06790926427908 0.55
1.07567357966892 0.55
1.0780092650946 0.55
1.07879253004147 0.55
1.08238609515997 0.55
1.08354837320105 0.555
1.08574135480815 0.555
1.08829570841889 0.56
1.09051165595912 0.565
1.0950463149416 0.565
1.09521045880803 0.57
1.09530042764255 0.575
1.09600008333333 0.575
1.10135514303828 0.58
1.10408921933085 0.585
1.1062818336163 0.585
1.11002444987775 0.59
1.1120019050643 0.595
1.11484593837535 0.595
1.122725871107 0.6
1.1310381294964 0.6
1.13237667401507 0.605
1.13480013550202 0.605
1.14012556903766 0.605
1.1406333870102 0.61
1.14192766759401 0.615
1.14639597690445 0.62
1.15068493150685 0.625
1.15238593745839 0.625
1.15249994299071 0.625
1.16362449139182 0.63
1.17492253066667 0.63
1.18231248538695 0.63
1.19492127342914 0.635
1.19637298805815 0.635
1.19726812371134 0.64
1.20735314612275 0.645
1.21225155665616 0.65
1.2147406733394 0.655
1.21481225524073 0.655
1.21952569460622 0.66
1.22072602459016 0.665
1.23001572645559 0.67
1.23203222174937 0.67
1.24193036418915 0.675
1.24391850355669 0.68
1.24817857345438 0.68
1.25060829396074 0.68
1.25384044644401 0.685
1.25871171754548 0.685
1.26674237879034 0.685
1.26676392362633 0.69
1.27864940493484 0.695
1.28797424051519 0.695
1.2939824537037 0.695
1.29656794907279 0.7
1.29796239126988 0.7
1.3026712823864 0.7
1.3200782680048 0.705
1.33157540394973 0.705
1.33220255231158 0.705
1.33641160949868 0.705
1.33985710290546 0.705
1.33985710290546 0.705
1.34871794871795 0.71
1.3582554517134 0.71
1.37442922374429 0.715
1.3799374169279 0.715
1.38752052545156 0.715
1.40692260044186 0.715
1.4089807836379 0.72
1.41257113295396 0.72
1.41351805205709 0.725
1.42471042471042 0.725
1.44185560968721 0.73
1.44358090769118 0.735
1.47236171656019 0.74
1.48862115127175 0.74
1.49384038564542 0.74
1.50178583766449 0.745
1.52545113547377 0.745
1.53564277711114 0.745
1.53852715382428 0.745
1.54213081369823 0.745
1.55835038463637 0.75
1.55918926465238 0.755
1.56451988330341 0.755
1.56869936305732 0.755
1.57169484171322 0.755
1.57484875757574 0.755
1.5977358490566 0.755
1.60845701715762 0.76
1.6242652558978 0.76
1.63279248505551 0.765
1.64080873950139 0.77
1.64080873950139 0.77
1.64108675721562 0.775
1.64236718124643 0.775
1.67110571390557 0.78
1.67686547895688 0.785
1.68555049261084 0.79
1.69003816004355 0.79
1.71669884169884 0.795
1.72125977191489 0.795
1.72204319459042 0.8
1.72204319459042 0.8
1.73183109348684 0.805
1.73856209150327 0.81
1.73856209150327 0.81
1.73918368677634 0.815
1.75845339422148 0.815
1.77120822622108 0.815
1.79557740835115 0.815
1.81331314821576 0.815
1.8551444152249 0.815
1.86378635862486 0.815
1.89261330525837 0.815
1.91658291457286 0.82
1.92810324315206 0.82
1.93075127435604 0.825
1.93410887289448 0.83
1.93563947751611 0.83
1.98 0.83
1.98569076599327 0.83
1.98716290322581 0.83
2.02559867877787 0.835
2.06079983318953 0.835
2.0616903978688 0.835
2.08114230210123 0.84
2.10430562506503 0.84
2.14128506724684 0.84
2.15144872695347 0.84
2.15962776477199 0.84
2.17139242783505 0.84
2.17407969324928 0.84
2.18582897340024 0.84
2.21248589695374 0.84
2.23233479986583 0.845
2.28119504685994 0.845
2.29514858874071 0.845
2.31623960398644 0.845
2.34471080114449 0.85
2.46124486821645 0.85
2.50234437478026 0.85
2.54240342380899 0.85
2.59858747420225 0.85
2.60248946551091 0.85
2.61857323015873 0.85
2.62017850814518 0.85
2.62958539119804 0.85
2.63205885867512 0.85
2.63340821990221 0.855
2.6553686252648 0.855
2.66458229114557 0.86
2.79926699418766 0.86
2.8071381887933 0.86
2.82906688687036 0.86
2.82966331175309 0.86
2.83612594557693 0.865
2.88133294587092 0.87
2.88201543165468 0.87
2.90796478506769 0.87
2.94802898915482 0.875
2.96410638960314 0.875
2.96837130131681 0.875
2.97033898305085 0.875
2.9846700389233 0.875
3.00147000419815 0.875
3.01240857763267 0.875
3.02698055370123 0.875
3.02796522572436 0.875
3.04781023725036 0.875
3.05076910005294 0.88
3.08855169946333 0.88
3.09554582215065 0.88
3.10440670666189 0.88
3.1658608845888 0.88
3.16732477788746 0.88
3.19796557120501 0.885
3.20588357080967 0.89
3.21909527275224 0.89
3.24057375228896 0.89
3.24900227472596 0.89
3.31825385824507 0.89
3.31930250549153 0.89
3.35261220958972 0.89
3.38899671814672 0.89
3.42154872018385 0.89
3.42460612279802 0.89
3.43574921404464 0.89
3.51051914985342 0.89
3.51402941176471 0.89
3.52601916260679 0.89
3.66475635876841 0.89
3.68221511814957 0.89
3.6825752327914 0.89
3.73810134157132 0.89
3.76100793874391 0.89
3.82801719274097 0.89
3.84675739698557 0.89
3.86223346793349 0.89
3.90545304601627 0.89
3.91859374874372 0.89
3.94198372520865 0.89
3.99458638405707 0.89
3.99716089124534 0.89
4.03676754382979 0.89
4.07066065681614 0.89
4.10749208251372 0.89
4.12235289013895 0.89
4.12453337484433 0.89
4.16345750857381 0.89
4.18424826090482 0.89
4.19038088453621 0.89
4.29361777564626 0.89
4.31226040515654 0.89
4.36831384767498 0.89
4.41424392352105 0.89
4.43021311674716 0.89
4.47712200075216 0.89
4.49304529765886 0.89
4.51111111111111 0.89
4.58961142857143 0.89
4.71868028301887 0.89
4.74943269848734 0.89
4.82877161088405 0.89
4.95103531050955 0.89
5.05653009004502 0.89
5.36861359372813 0.89
5.41371495271868 0.89
5.50454139216618 0.89
5.61927910150697 0.89
5.64347023911187 0.89
5.66520140293961 0.89
5.67138455008489 0.89
5.7172490862423 0.89
5.74192375210692 0.89
5.7944063771518 0.89
5.84623600552868 0.89
5.92644768932018 0.89
6.06668544017296 0.89
6.16879451252895 0.89
6.20837482953296 0.89
6.21431991732711 0.89
6.21431991732711 0.89
6.24325325872284 0.895
6.2557923503861 0.895
6.26902996375352 0.895
6.32113694762684 0.895
6.33118291377718 0.895
6.39088494835219 0.895
6.60209476439791 0.895
6.62922038154248 0.895
6.67407481481481 0.895
6.7505980013035 0.895
6.79605413231066 0.895
6.84771066389964 0.895
6.87053019061051 0.895
6.92431147594671 0.895
6.92431147594671 0.895
7.02771102642041 0.895
7.0703344 0.895
7.07247365369606 0.895
7.24802053835804 0.895
7.24802053835804 0.895
7.29669833267304 0.9
7.34546383326968 0.905
7.50535055749521 0.905
7.52929229188078 0.905
7.95458560411311 0.905
8.07105606258149 0.905
8.15290387743683 0.905
8.23619968331987 0.905
8.26659608801251 0.905
8.52526693447599 0.91
8.57800134728033 0.91
8.95905296957672 0.91
9.06703997837469 0.91
9.138502118613 0.91
9.14117647058823 0.91
9.19497104059463 0.915
9.27184124472136 0.915
9.27184124472136 0.915
9.44323963397198 0.92
9.46920699921789 0.92
9.47410512820513 0.92
9.5886230043499 0.92
9.5899465506836 0.92
9.62998818609514 0.92
9.70272713032778 0.92
10.0047281323877 0.92
11.1268536574074 0.92
11.3147221376039 0.92
11.3147221376039 0.92
12.1358851583113 0.925
12.448430941704 0.925
12.5156926927891 0.925
12.9678954720088 0.93
13.6333176026428 0.93
13.6333176026428 0.93
14.553184921067 0.935
15.1614041461007 0.935
17.1414161392405 0.935
17.3913712374582 0.94
18.5768076593312 0.945
21.0440566082425 0.95
21.2749365845675 0.95
22.0776293759513 0.95
23.2808918338109 0.95
29.5187319646208 0.955
31.2101575999066 0.955
34.750181438687 0.96
38.0563524590164 0.96
42.3937714006196 0.96
43.3539377688299 0.965
49.7418770009789 0.965
50.9882411764706 0.97
52.0018547479126 0.97
52.4724478326985 0.97
117.0001 0.975
234.0001 0.98
329.000100000299 0.98
489.0001 0.985
591.0002 0.99
1020.0001 0.995
1375.0003 0.995
1531.0005 0.995
1936.0005 0.995
2237.0001 1
2308.0006 1
3073.0012 1
4211.0003 1
9161.001 1
12283.0014000112 1
25165.0063 1
};
\addlegendentry{Decentralized}
\addplot [semithick, forestgreen4416044]
table {%
1 0
1 0
1 0
1 0
1 0
1 0
1 0
1 0
1 0
1 0
1 0
1 0
1 0
1 0
1 0
1 0
1 0
1 0
1 0
1 0
1 0
1 0
1 0
1 0
1 0
1 0
1 0
1 0
1 0
1 0
1 0
1 0
1 0
1 0
1 0
1 0
1 0
1 0
1 0
1 0
1 0
1 0
1 0
1 0
1 0
1 0
1 0
1 0
1 0
1 0
1 0
1 0
1 0
1 0
1 0
1 0
1 0
1 0
1 0
1 0
1 0
1 0
1 0
1 0
1 0
1 0
1 0
1 0
1 0
1 0
1 0
1 0
1 0
1 0
1 0
1 0
1 0
1 0
1 0
1 0
1 0
1 0
1 0
1 0
1 0
1 0
1 0
1 0
1 0
1 0
1 0
1 0
1 0
1 0
1 0
1 0
1 0
1 0
1 0
1 0
1 0
1 0
1 0
1 0
1 0
1 0
1 0
1 0
1 0
1 0
1 0
1 0
1 0
1 0
1 0
1 0
1 0
1 0
1 0
1 0
1 0
1 0
1 0
1 0
1 0
1 0
1 0
1 0
1 0
1 0
1 0
1 0
1 0
1 0
1 0
1 0
1 0
1 0
1 0
1 0
1 0
1 0
1 0
1 0
1 0
1 0
1 0
1 0
1 0
1 0
1 0
1 0
1 0
1 0
1 0
1 0
1 0
1 0
1 0
1 0
1 0
1 0
1 0
1 0
1 0
1 0
1 0
1 0
1 0
1 0
1 0
1 0
1 0
1 0
1 0
1 0
1 0
1 0
1 0
1 0
1 0
1 0
1 0
1 0
1 0
1 0
1 0
1 0
1 0
1 0
1 0
1 0
1 0
1 0
1 0
1 0
1 0
1 0
1 0
1 0
1 0
1 0
1 0
1 0
1 0
1 0
1 0
1 0
1 0
1 0
1 0
1 0
1 0
1 0
1 0
1 0
1 0
1 0
1 0
1 0
1 0
1 0
1 0
1 0
1 0
1 0
1 0
1 0
1 0
1 0
1 0
1 0
1 0
1 0
1 0
1 0
1 0
1 0
1.00140789247829 0.625
1.00279460038502 0.625
1.00379162604268 0.625
1.00853807283499 0.625
1.00896905829596 0.63
1.00979586684715 0.63
1.01107601600337 0.635
1.01116389548694 0.64
1.01249132546842 0.645
1.01298945298207 0.645
1.01303538175047 0.645
1.01790419566246 0.645
1.02144485495255 0.645
1.02281144781145 0.645
1.02288985966346 0.65
1.02734103769525 0.65
1.02869130844511 0.655
1.02934169278997 0.66
1.03779958826497 0.665
1.04499430771806 0.67
1.04892086330935 0.67
1.05330862890418 0.67
1.05466834303175 0.675
1.05632658768572 0.68
1.05838521033767 0.685
1.06083011079731 0.69
1.06174603174603 0.69
1.06790926427908 0.695
1.07567357966892 0.7
1.0780092650946 0.705
1.07879253004147 0.71
1.08238609515997 0.715
1.08354837320105 0.715
1.08574135480815 0.72
1.08829570841889 0.72
1.09051165595912 0.72
1.0950463149416 0.725
1.09521045880803 0.725
1.09530042764255 0.725
1.09600008333333 0.73
1.10135514303828 0.73
1.10408921933085 0.73
1.1062818336163 0.735
1.11002444987775 0.735
1.1120019050643 0.735
1.11484593837535 0.74
1.122725871107 0.74
1.1310381294964 0.745
1.13237667401507 0.745
1.13480013550202 0.75
1.14012556903766 0.755
1.1406333870102 0.755
1.14192766759401 0.755
1.14639597690445 0.755
1.15068493150685 0.755
1.15238593745839 0.76
1.15249994299071 0.765
1.16362449139182 0.765
1.17492253066667 0.765
1.18231248538695 0.77
1.19492127342914 0.77
1.19637298805815 0.775
1.19726812371134 0.775
1.20735314612275 0.775
1.21225155665616 0.775
1.2147406733394 0.775
1.21481225524073 0.78
1.21952569460622 0.78
1.22072602459016 0.78
1.23001572645559 0.78
1.23203222174937 0.785
1.24193036418915 0.785
1.24391850355669 0.785
1.24817857345438 0.79
1.25060829396074 0.795
1.25384044644401 0.795
1.25871171754548 0.8
1.26674237879034 0.8
1.26676392362633 0.8
1.27864940493484 0.8
1.28797424051519 0.805
1.2939824537037 0.81
1.29656794907279 0.81
1.29796239126988 0.815
1.3026712823864 0.82
1.3200782680048 0.82
1.33157540394973 0.825
1.33220255231158 0.83
1.33641160949868 0.835
1.33985710290546 0.84
1.33985710290546 0.84
1.34871794871795 0.845
1.3582554517134 0.85
1.37442922374429 0.85
1.3799374169279 0.855
1.38752052545156 0.855
1.40692260044186 0.86
1.4089807836379 0.86
1.41257113295396 0.865
1.41351805205709 0.865
1.42471042471042 0.87
1.44185560968721 0.87
1.44358090769118 0.87
1.47236171656019 0.87
1.48862115127175 0.875
1.49384038564542 0.88
1.50178583766449 0.88
1.52545113547377 0.885
1.53564277711114 0.885
1.53852715382428 0.885
1.54213081369823 0.89
1.55835038463637 0.89
1.55918926465238 0.89
1.56451988330341 0.895
1.56869936305732 0.895
1.57169484171322 0.895
1.57484875757574 0.895
1.5977358490566 0.895
1.60845701715762 0.895
1.6242652558978 0.9
1.63279248505551 0.9
1.64080873950139 0.9
1.64080873950139 0.9
1.64108675721562 0.905
1.64236718124643 0.905
1.67110571390557 0.905
1.67686547895688 0.905
1.68555049261084 0.905
1.69003816004355 0.905
1.71669884169884 0.905
1.72125977191489 0.91
1.72204319459042 0.91
1.72204319459042 0.91
1.73183109348684 0.915
1.73856209150327 0.915
1.73856209150327 0.915
1.73918368677634 0.92
1.75845339422148 0.92
1.77120822622108 0.92
1.79557740835115 0.925
1.81331314821576 0.93
1.8551444152249 0.93
1.86378635862486 0.93
1.89261330525837 0.935
1.91658291457286 0.935
1.92810324315206 0.94
1.93075127435604 0.94
1.93410887289448 0.94
1.93563947751611 0.945
1.98 0.945
1.98569076599327 0.945
1.98716290322581 0.945
2.02559867877787 0.945
2.06079983318953 0.945
2.0616903978688 0.945
2.08114230210123 0.945
2.10430562506503 0.945
2.14128506724684 0.945
2.15144872695347 0.945
2.15962776477199 0.945
2.17139242783505 0.945
2.17407969324928 0.945
2.18582897340024 0.945
2.21248589695374 0.945
2.23233479986583 0.945
2.28119504685994 0.945
2.29514858874071 0.945
2.31623960398644 0.945
2.34471080114449 0.945
2.46124486821645 0.945
2.50234437478026 0.945
2.54240342380899 0.945
2.59858747420225 0.95
2.60248946551091 0.95
2.61857323015873 0.95
2.62017850814518 0.95
2.62958539119804 0.95
2.63205885867512 0.95
2.63340821990221 0.95
2.6553686252648 0.955
2.66458229114557 0.955
2.79926699418766 0.96
2.8071381887933 0.96
2.82906688687036 0.96
2.82966331175309 0.965
2.83612594557693 0.965
2.88133294587092 0.965
2.88201543165468 0.965
2.90796478506769 0.965
2.94802898915482 0.965
2.96410638960314 0.965
2.96837130131681 0.965
2.97033898305085 0.965
2.9846700389233 0.965
3.00147000419815 0.965
3.01240857763267 0.965
3.02698055370123 0.97
3.02796522572436 0.97
3.04781023725036 0.97
3.05076910005294 0.97
3.08855169946333 0.97
3.09554582215065 0.97
3.10440670666189 0.97
3.1658608845888 0.97
3.16732477788746 0.97
3.19796557120501 0.97
3.20588357080967 0.97
3.21909527275224 0.97
3.24057375228896 0.97
3.24900227472596 0.975
3.31825385824507 0.975
3.31930250549153 0.975
3.35261220958972 0.975
3.38899671814672 0.975
3.42154872018385 0.975
3.42460612279802 0.975
3.43574921404464 0.975
3.51051914985342 0.975
3.51402941176471 0.975
3.52601916260679 0.975
3.66475635876841 0.975
3.68221511814957 0.975
3.6825752327914 0.975
3.73810134157132 0.975
3.76100793874391 0.975
3.82801719274097 0.975
3.84675739698557 0.975
3.86223346793349 0.975
3.90545304601627 0.975
3.91859374874372 0.975
3.94198372520865 0.975
3.99458638405707 0.975
3.99716089124534 0.975
4.03676754382979 0.975
4.07066065681614 0.975
4.10749208251372 0.975
4.12235289013895 0.975
4.12453337484433 0.975
4.16345750857381 0.975
4.18424826090482 0.975
4.19038088453621 0.975
4.29361777564626 0.975
4.31226040515654 0.975
4.36831384767498 0.975
4.41424392352105 0.975
4.43021311674716 0.975
4.47712200075216 0.975
4.49304529765886 0.975
4.51111111111111 0.975
4.58961142857143 0.98
4.71868028301887 0.98
4.74943269848734 0.98
4.82877161088405 0.98
4.95103531050955 0.98
5.05653009004502 0.98
5.36861359372813 0.98
5.41371495271868 0.98
5.50454139216618 0.98
5.61927910150697 0.98
5.64347023911187 0.98
5.66520140293961 0.98
5.67138455008489 0.98
5.7172490862423 0.98
5.74192375210692 0.98
5.7944063771518 0.98
5.84623600552868 0.98
5.92644768932018 0.98
6.06668544017296 0.98
6.16879451252895 0.98
6.20837482953296 0.98
6.21431991732711 0.98
6.21431991732711 0.98
6.24325325872284 0.985
6.2557923503861 0.985
6.26902996375352 0.985
6.32113694762684 0.985
6.33118291377718 0.985
6.39088494835219 0.985
6.60209476439791 0.985
6.62922038154248 0.985
6.67407481481481 0.985
6.7505980013035 0.985
6.79605413231066 0.985
6.84771066389964 0.985
6.87053019061051 0.985
6.92431147594671 0.985
6.92431147594671 0.985
7.02771102642041 0.99
7.0703344 0.99
7.07247365369606 0.99
7.24802053835804 0.99
7.24802053835804 0.99
7.29669833267304 0.99
7.34546383326968 0.99
7.50535055749521 0.99
7.52929229188078 0.99
7.95458560411311 0.99
8.07105606258149 0.99
8.15290387743683 0.99
8.23619968331987 0.99
8.26659608801251 0.99
8.52526693447599 0.99
8.57800134728033 0.99
8.95905296957672 0.99
9.06703997837469 0.99
9.138502118613 0.99
9.14117647058823 0.99
9.19497104059463 0.99
9.27184124472136 0.99
9.27184124472136 0.99
9.44323963397198 0.99
9.46920699921789 0.99
9.47410512820513 0.99
9.5886230043499 0.99
9.5899465506836 0.99
9.62998818609514 0.99
9.70272713032778 0.99
10.0047281323877 0.99
11.1268536574074 0.995
11.3147221376039 0.995
11.3147221376039 0.995
12.1358851583113 0.995
12.448430941704 0.995
12.5156926927891 0.995
12.9678954720088 0.995
13.6333176026428 0.995
13.6333176026428 0.995
14.553184921067 0.995
15.1614041461007 0.995
17.1414161392405 0.995
17.3913712374582 0.995
18.5768076593312 0.995
21.0440566082425 0.995
21.2749365845675 0.995
22.0776293759513 0.995
23.2808918338109 0.995
29.5187319646208 0.995
31.2101575999066 1
34.750181438687 1
38.0563524590164 1
42.3937714006196 1
43.3539377688299 1
49.7418770009789 1
50.9882411764706 1
52.0018547479126 1
52.4724478326985 1
117.0001 1
234.0001 1
329.000100000299 1
489.0001 1
591.0002 1
1020.0001 1
1375.0003 1
1531.0005 1
1936.0005 1
2237.0001 1
2308.0006 1
3073.0012 1
4211.0003 1
9161.001 1
12283.0014000112 1
25165.0063 1
};
\addlegendentry{Centralized}
\end{axis}

\end{tikzpicture}